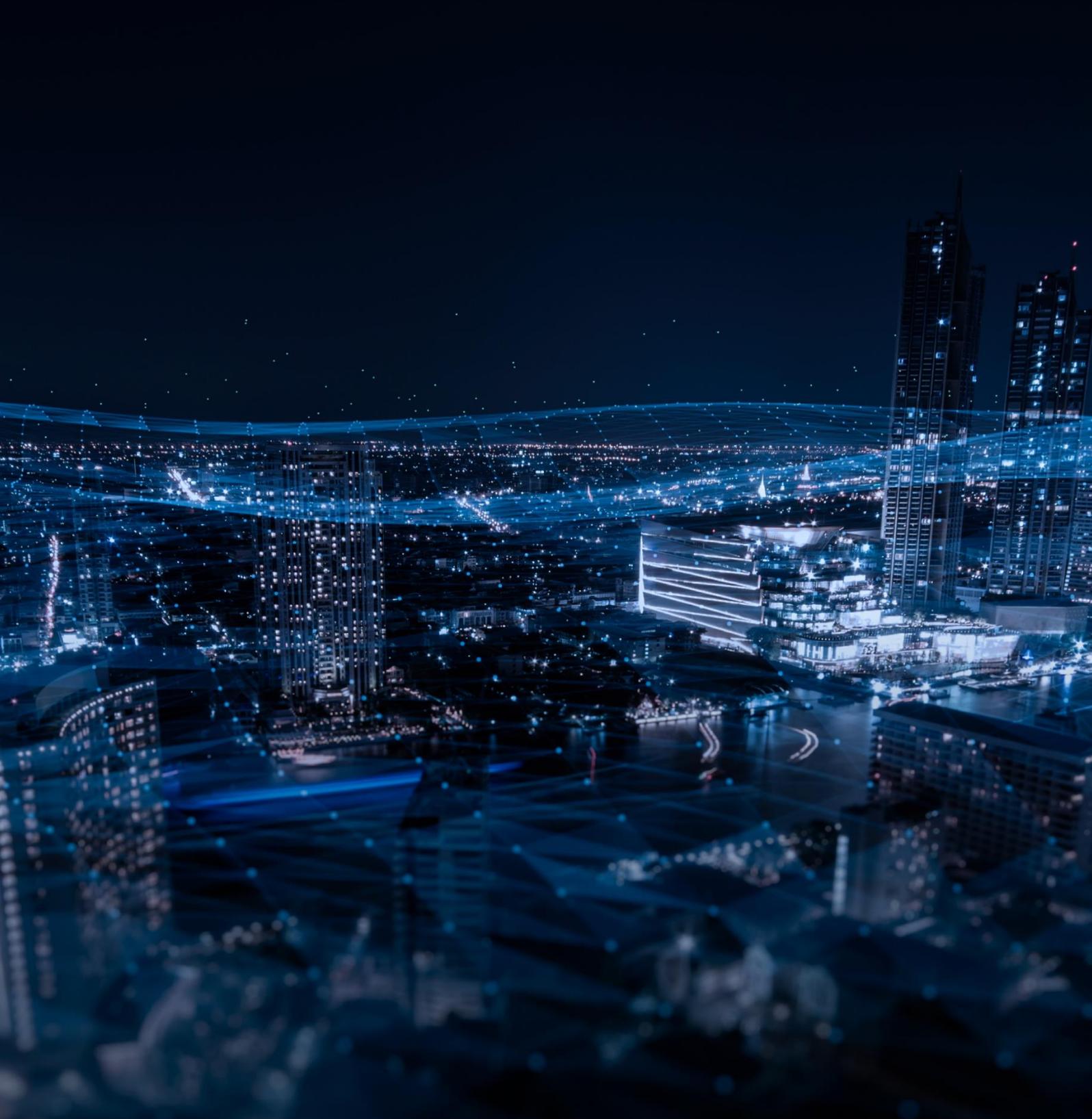

# White Paper on 6G BSS Technologies

Jointly released by:

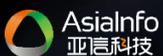 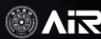 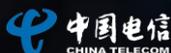 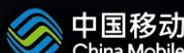 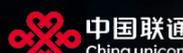 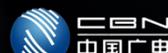 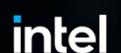

# Authors

AsiaInfo Technologies

Institute for AI Industry Research, Tsinghua University

China Telecom Research Institute

China Mobile Information Technology Center

China Unicom Software Research Institute

China Broadcasting Network Mobile Network Co., Ltd.

Intel (China) Co., Ltd.

## Cite This White Paper

# Contents

# List of Figures

# List of Tables

# I. Preface

With the evolution of communication technology from 1G to 5G, operators have also experienced changes from the era of telecom services to consumer Internet and then industrial Internet. The diversification and specialization of the information and communication business offer operators new avenues of development. As the digital economy gains momentum in boosting global economic growth, traditional CSP (communication service provider) operators are prioritizing integrated development of ICT (Information and Communication Technology) infrastructure and business, upgrading cloud network infrastructure, creating a new business model of cloud-network integration, and actively promoting the digital and intelligent transformation of the whole industry to empower the development of global digital economy [1].

In the 1990s, the commercialization of the Internet began a new era of digital economy, known as Web1.0, characterized by the "Internet of information" and personal computers. 1G/2G digital technologies just began to replace analog technology to provide voice communication, and mobile communication and the Internet were completely independent from each other. In 2004, the Web2.0 era, or the era of "interactive Internet" began, allowing users to not only consume information, but publish their own content on the Internet. The 3G/4G mobile communication technologies enabled the public to share text and pictures through smart mobile terminals, leading to the rise of mobile Internet with better experience and bringing great commercial benefits to operators. This ultimately led to the consumer Internet which constitutes the foundation of digital economy. With 5G commercialization, operators are accelerating their transformation into digital service operators, hoping to seize the opportunities brought by digital economy as builders of digital infrastructure and enablers of industrial digital and intelligent transformation. According to *G20 Digital Economy Development and Cooperation Initiative* adopted by G20 leaders at the G20 Summit in September 2016, the digital economy refers to a broad range of economic activities that use digitized information and knowledge as the key factor of production, modern information networks as the important activity space, and information and communication technology as an important driver for efficiency-enhancing and economic structural optimization[2]. On September 4, 2019, the United Nations Conference on Trade and Development (UNCTAD) released its first *Digital Economy Report*, stating that an entirely new "data value chain" has evolved, and businesses that build digital platforms have a major advantage in the data-driven economy [3]. With the advantages of the Internet of Everything in the 5G network, operators are promoting cloud-network collaboration and computing and network convergence, while establishing computing-aware networks based on their cloud-network infrastructure resources. The operators will also promote the digital upgrading of thousands of industries through a strong government & enterprise customer base and the ever-increasing integrated solution capability, which will create a medium- and long-term expansion engine for operators and lay the foundation for the sustainable and high-quality business development. In addition, the combination of 5G and artificial intelligence (AI) has been regarded by the industry as the latest generation of general purpose technologies, which will boost productivity and empower vertical industries, and accelerate the development of AI in the communication ecology domain over the next decade [4].

In the evolution from 5G to 6G, the development of Internet and IT (Information Technology) provides strong support for the informatization of 6G-oriented CT (Communication Technology). Based on this, DOICT (Data, Operation, Information and communication technologies) convergence will be the core of digital technology innovation in the future and the evolution direction of digital information infrastructure. Web3.0 proposed in 2014, which represents the next generation of a decentralized Internet, aims to realize "those who create will own" through decentralized technologies such as blockchain so as to break up the current monopoly of Internet companies and reshape the Internet value chain. Users own the content and data they created, and the value generated thereof can also be distributed as per the agreement between the platform and users. Therefore, Web3.0 is also called the "Internet of Value" [5]. The metaverse based on Web3.0 is a virtual world that is linked and created by means of technology, mapped and interacted with the real world, and a digital living space with a new social system. It can be said that it is a new type of Internet application and social form that integrates multiple new technologies. In the ideal metaverse, the digital world will be as important as the real world. Inside the metaverse, users can enter the digital world at any time by utilizing digital avatars and confirming their identity through the use of Internet tokens, and they also have more extensive control over their personal data. Although communication operators have established their position in the digital economy by developing 4G and 5G networks in the Web2.0 era, and won corresponding opportunities in consumer business and enterprise-level business respectively, they are plagued by bottlenecks in the traditional consumer business and diversified demands of enterprise-level business, which cannot be settled by a single communication technology in a closed loop. In this context, communication operators hope to develop 6G network technology that is more efficient, cost-effective, and can support novel business models. This is in response to consumers' constant demand for better experience, the vertical industry's need for high-quality and flexible radio networks, and the vision of innovative development through Web3.0 business model. The metaverse has huge demands for computing and network resources. The massive real-time information interaction and immersive experience of the metaverse on the infrastructure need to be based on the continuous improvement of communication technology and computing power. As the main provider of computing and network services, operators can undoubtedly enhance the perceptual and intelligent capacities of 6G networks to effectively connect computing power services to support new application, and establish a crucial position in the metaverse by taking the initiative to become service providers of metaverse infrastructure [6]. In addition, in the 6G era, operators need to further integrate the relevant DOICT to realize the business support capability of computing and network convergence, and develop diversified value-added business through "connection + computing power + capability" based on their profound network technology and operation technology, so as to better meet the development of next-generation Internet applications.

To sum up, in developing 6G networks and continuously transforming into digital service providers, communication operators also pay attention to new business models brought about by the development of the next-generation Internet and IT while focusing on the intergenerational research of communication technology. They should also upgrade from "network-centric" to "business and service centric" and "customer-centric" according to the basic laws of digital economy. The upgrade of IT support system, which carries business development and operation, will also be an important part of its own transformation. The traditional IT support system of operators includes three subsystems: business support system (BSS), operation support system (OSS) and management support system (MSS), which are interrelated and bear different responsibilities. However, with the improvement of 6G network capability, the market demand is more extensive and the competition is more intense, which requires to further strengthen BSS and OSS integration regarding basic 6G computing and network services, to improve the operational efficiency and benefits of customers by vertically connecting the digital intelligence support capabilities on both sides of supply and demand, and to further integrate BSS and MSS so as to horizontally connect the whole business process from strategy to implementation and realize business innovation.

This White Paper will focus on business requirements of operators in the 6G era, as well as how they evolve and upgrade their digital BSS to realize sustainable business innovation.

# II. Evolution of CSP Business and BSS

BSS refers to a CSP business supporting system, which usually consists of a set of software and hardware components that support the operation, management and maintenance activities of telecom service providers and provide customer-oriented services [7,8]. The main modules of BSS include customer management, order management, product management, tariff management, expense management, etc., which can help operators automate and optimize business processes, improve business efficiency and customer service quality, and reduce operation costs and risks. In the digital economy era, operators have been integrating Internet-related technologies with the development of communication technology to expand their business domain and promote the synchronous evolution of BSS, as shown in *Figure 2-1*. Due to the intergenerational upgrade of the communication technology, the main business of operators in the same period gradually moves from communication service to digital service oriented to Internet of Everything. BSS has also evolved into the cloudified architecture, thus more effectively supporting the business expansion and innovation of operators.

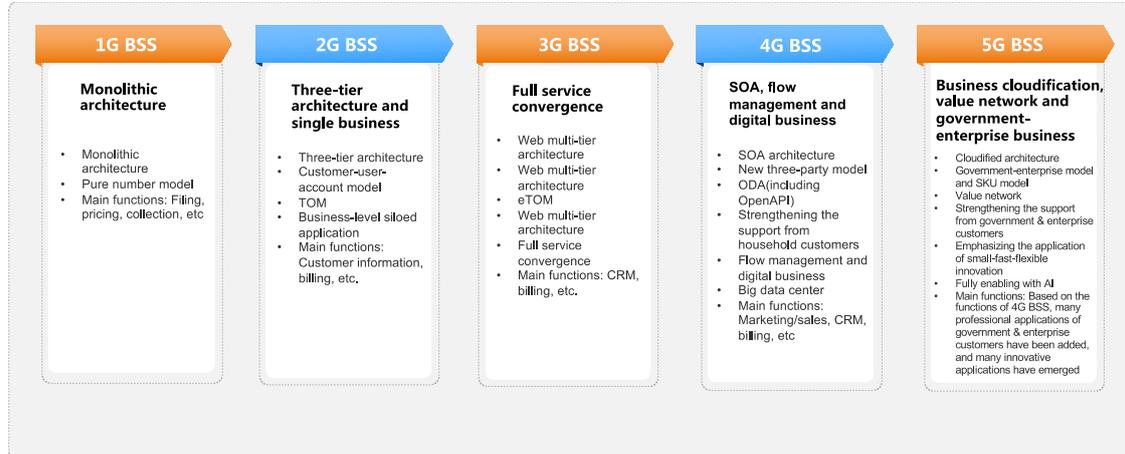

**Figure 2-1 Intergenerational Characteristics of BSS of Operators**

## 2.1 BSS Characteristics in Communication Generation

### 1G BSS: Monolithic architecture

In 1973, Motorola developed the first simulated handheld terminal, marking the beginning of mobile communication business which later expanded to Japan, Europe, the United States and China successively [9]. In the 1G era, devices and networks at their primary stage of development had limited capacities and could only support voice communication. Devices and handheld terminals were too expensive to be widely used for civilian purpose. Therefore, the business system mainly provided support for phone numbers based on voice business, including basic functions such as number management, number selection, monthly billing, opening of selected number, number billing and fee collection.

The 1st generation BSS back then was still in the initial phase in terms of capability support, so the single server solution was enough to meet the demand. At that time, the computer system was not advanced, and the application systems were all independent of each other, with no layered service or modern databases and Web servers. Application codes were usually written in C, Fortran, COBOL, dBASE (and compatible products), compiled into single executable files, or run by script loading.

## 2G BSS: Three-tier architecture and monolithic service

In 1982, the GSM specifications were formulated for mobile communication, and the 2G era began. After the 1G GSM commercialization, and the inter-generational development of 2.5G GPRS and 2.75G EDGE, the digitization and informatization of communication technology and the large-scale chip integration technology addressed the problems of expensive device, roaming and insufficient spectrum. Mobile communication became popular among people with a wider range of business available, such as short messages and simple Internet.

Telecommunication Management Forum (TM Forum), established in 1988, is a worldwide organization that provides strategic suggestions and implementation plans for telecommunications operation and management. It has been focusing on the technical system and operational efficiency of BSS and the revenue capability of operators for many years. TM Forum put forward the Telecom Operations Map (TOM) model of BSS in the mid-1990s, which covers business processes such as customer relationship management, order management, product management, tariff management and expense management [10]. The 2G BSS helped operators realize business process optimization and automation by reference to the business flow chart and related standards in TOM.

This generation of BSS was still positioned to meet the management needs of operators themselves, but has begun to seek breakthroughs and innovations in customer service experience. Entering the 2G era, mobile communication business thrived, offering more sophisticated business support functions such as SMS, paging, MMS and simple Internet access.

The functional design of BSS has the following characteristics:

- BSS began to split into business, billing and customer service subsystems, in which the business system was responsible for completing the sales, changes and payment of mobile phones, numbers and services, the billing system was responsible for completing the data collection, measurement, pricing and billing, and the customer service system was responsible for supporting customer service by separating the traffic from the system.

- A new cooperation paradigm was introduced. In the 2G era, partners began to join the service, but it was mainly value-added service provided based on language and short messages, and BSS mainly provided settlement for value-added service providers.

- A user-centered customer-user-account model [11] was initiated in terms of system design. With the prosperity of business, there were a surge of users, and the characterization of users' attributes in various dimensions was getting more complex, such as the merging relationship between users and customers and the expression of payment relationship of accounts. Therefore, telecom operators put forward a customer-user-account model to adapt to the complicated payment scenarios and the merging management of users according to customers at that time.

In terms of technical support, BSS has the following characteristics:

- **Three-tier architecture:** In the 2G era, the design philosophy of "high cohesion, low coupling" in software architecture emerged, aiming to establish a clear division of labor, promote reuse of functions and clarify business logic, so as to speed up development, improve development efficiency and facilitate maintenance. For this reason, the communication business support system began to adopt the three-tier application architecture [presentation layer (UI), business logic layer (BLL) and data access layer (DAL), accessible to each other through interfaces], which greatly promoted the business system standards and laid a foundation for large-scale development of the system, as well as a good foundation for the further differentiation of the subsequent technical architecture and the formation of standard components. Personal computers appeared in this period, and the launch of desktop systems such as windows/xwindows provided physical and technical support for the rapid development of BSS.

- **Business-level siloed application:** In the 2G era, no systematic business operation support theory was formed in the communication field. In the face of rapid business development, the iteration speed of business system was obviously lagging behind. During the business development, the business operation support system was divided into business, billing, customer service and other systems, each of which was based on a three-tier architecture system, and a separate C/S version was established through a middleware system. The data, services and interfaces were independent of each other, and interface communication was adopted between systems [12].

### 3G BSS: Service convergence

The 3G network integrated radio communication with Internet and other multimedia communication technologies, enabling the provision of voice and high-speed data services simultaneously on a global scale. During this era, smart phones became increasingly prevalent, and information services and multimedia services emerged one after another, such as browsing web pages on mobile devices, sending and receiving emails, making video calls and watching live streams. Various Apps have been constantly updated and iterated, and the multimedia era has arrived. In order to respond to the market in time and effectively organize resources to form products/services, communication operators began to provide converged services. The business system quickly configured the combination of different products and tariffs to form new products so as to meet the flexible changes of market strategies. In this context, the product-commodity model evolved, including product specifications, attribute specifications, attribute values, commodity specifications, tariff specifications, among others. Product specifications and commodity specifications can include different attribute specifications and be configured with different attribute values. Commodities may incorporate different products, as well as different tariff specifications as their components. The product-commodity model is the core of the business operation system.

With the development of the telecom industry and advancement in technology, TOM framework has gradually shown weaknesses, e.g. its limited management scope and difficulty in supporting emerging technologies. Therefore, TM Forum put forward the eTOM (enhanced Telecom Operations Map) framework in 2001 as an improved and upgraded version of the TOM framework. The eTOM framework is designed to be more comprehensive, flexible and modern, extending and updating the TOM framework. The eTOM framework covers all business activities of telecom operators, including customer management, business development, operation management, resource management, quality management, among others, and it can support various traditional and emerging technologies.

With the TOM framework as the predecessor and basis, the eTOM framework is the evolution and upgrade of the TOM framework, providing more best practices and standardized business processes. The eTOM framework has become an international standard of telecom service management, and has been widely used in the management practice of telecom operators. Thanks to the eTOM framework, the 3G BSS can provide telecom operators with more comprehensive and in-depth business process management and optimization.

At this phase, with the increasing competition in the communication market, operators began to turn to customer-centric, committed to providing customers with better pre-sale, sale and after-sale services. In addition, in terms of cooperation paradigm, BSS began to provide comprehensive management of upstream and downstream partners, including partner life cycle management, partner product management, partner settlement management, etc.

At this phase, the design of BSS has the following characteristics:

- More comprehensive functional coverage: Supporting voice, SMS, MMS, data and other services, and providing management of marketing, channels, customers, customer service, orders, products/commodities, convergent billing, resources and partners.

- Customer-centric: Although a customer-user-account model was established in the 2G era, the business was centered on commodities. In the 3G era, due to the intensified competition, it is necessary to adjust the market strategy in real time according to the market situation and competitors' strategies to complete various combined commodities, and the marketing business model has changed from commodity-centered to "customer-centric", driven by the need to meet customers' demands.

Technically, the multi-tier architecture was introduced to support the system design reform of BSS: In the 3G era, the application framework shifted from CS to BS, accompanied by the transition from Web1.0 to Web2.0. This led to the emergence of MVC application architecture, in which M stands for model, V stands for view, and C stands for control. With business segmented and services being reused, control, model and view layers are subdivided into more layers, and RPC and Restful API open technologies were developed alongside MVC.

### 4G BSS: SOA, flow management and digital business

With the enhancement of 4G support, more business potentials have been released, including mobile payment, App store, mobile Internet, DiDi ride-hailing service, Meituan Takeaway, mobile e-commerce, etc. Various new businesses have mushroomed, changing people's lives completely by offering a more intelligent lifestyle.

At this phase, operators have shifted from providing customer communication and network connection services to meeting customers' comprehensive information service needs. In order to better seize the booming development opportunities of mobile Internet and cope with the challenge of the rising OTT vendors, operators have begun to cooperate with mainstream Internet service providers in depth. Therefore, BSS has higher requirements in supporting the openness and automation of value chain cooperation.

At this phase, the functional design of BSS has the following characteristics:

- **Flow management capability is introduced:** It serves as the key to adapt to the development law of mobile Internet, grasp the development opportunity of mobile Internet and change the operators' role as "plumber" in the Internet era. Flow management has promoted the transformation of operators to information service providers and fundamentally reshaped the value creation approach of telecom operators.

- **Digital business innovation:** In the digital age where data has become a valuable asset, it's natural for the operators to extend their business via datalization, i.e., using the collected data for the business or the product itself. It includes two levels, data intelligence and data innovation. As for the data intelligence, big data technology is used to enhance product experience, such as recommending system and credit rating. As for the data innovation, the accumulated data is mainly used to develop new business. The essence of digital business is the productization, commercialization and valuation of data.

- **New three-party model** [13]: With the monetization of ebb tide services, service owners provide services for service users by using their own surplus service value, and more services can be monetized by service users paying directly or through third party payment service. The three-party relationship involved in this monetization is different from the previous customer-user-account model in the 2G era, forming a new three-party model relationship, that is, the many-to-many relationship among the owner, the user and the payer. Among them, the owner can own the products of multiple operators, and these products can be used by users different from the owner. The bills from different owners can be paid by the users themselves or through different payers, and users can even pay part of the bills from different owners through different payers, thus forming a very complicated and flexible many-to-many relationship.

Technically, BSS has introduced the current new computer application development mode, presenting the following characteristics:

- **SOA architecture:** In order to support a large number of innovative services in the 4G era, BSS has adopted the SOA architecture, and introduced technologies such as enterprise service bus (ESB), microservice centralization, and business middle platform. The essence of the SOA is to improve service reuse, realize flexible and standardized management and rapid self-adaptive elastic scalability, accelerate rapid business iteration and meet market demand quickly.

- **Building a big data center:** Mobile Internet services have brought massive unstructured data to operators, including behavior data, access data, etc. In order to release the data value, explore business opportunities, discover users' consumption habits, and reasonably recommend more services, operators have introduced big data centers, and then concentrated on analysis and processing according to business needs.

## 5G BSS: Service cloudification and value network

In the 5G era, the communication network infrastructure is moving away from specialized hardware platform towards the widespread adoption of basic hardware architectures such as x86, ARM and others. The commercialization of the 5G network technology has further enriched the application scenarios, creating application foundations including industrial Internet, autonomous driving, AR/VR, metaverse and so on. At this phase, while fully meeting the needs of individual customers, operators began to focus on the information service demands of government/enterprise customers. In addition, from the perspective of cooperation paradigm, the partners at this phase are more diverse, including vertical industry partners, various ISPs, etc. Therefore, BSS business has also changed from supporting the cooperation paradigm centered on operators to supporting the operation partner network paradigm.

At this phase, the functional design of BSS has the following characteristics:

- **Government-enterprise model and SKU model:** In order to better support the 2B services, BSS began to strengthen the government-enterprise model. In addition, due to the highly integrated and mixed characteristics of Internet-based and telecom-based commodities, telecom operators have also added SKU-related models (including related complete sets of models, such as SPU) to BSS to track inventory and determine the sales price and inventory level of the goods.

- **Strengthening the marketing and service of government & enterprise customers:** There are a large number of ToB-oriented services in the 5G era, and indicators such as latency, bandwidth, connections and reliability are more personalized. Therefore, starting with the industry user business, the 5G era features the focus on industry user ecology and promotes BSS innovation.

- **Emphasizing the application of small-fast-flexible innovation:** 5G has solved various network reliability problems for industry users, and has actually built scenario ecologies for users through combination with matured cloud technology and the emerging edge cloud. In these scenario ecologies, many mature Internet and mobile Internet capabilities and services have been or will be integrated, so in order to meet the urgent needs of industry users, developers can use these scenario ecologies to quickly build industry application capabilities.

- **Value network** [14]**:** Value network is the expansion and promotion of the value chain. The value network is to reconstruct the original value chain around the customer value, and it is a network of connections and exchanges of multiple value chains in multiple links through the interaction between different levels and different subjects in the network. The network formed by these relationships will generate network effects, and individuals or organizations at each network node can create or gain more value from this aggregation. The functional design of the 5G BSS has taken the support for the value network into consideration, and has made various explorations in maintaining the customer ecology.

The technical system of cloud computing has been incorporated into the system architecture of BSS, specifically including:

- **Business cloudification:** Including two characteristics, i.e. cloud-edge integration and cloud native. In order to address the timeliness and security of the 5G industrial control, cloud services have evolved into cloud and edge cloud, and in order to meet businesses' demand for elasticity and flexibility, a large number of cloud native technologies such as Docker technology and K8s dynamic scalability are introduced.

- **Full intelligence injection:** Cloud networks convergence and scenario ecology in the 5G era result in a very complicated system from the perspectives of technology, business, operation and management. From the perspective of underlying technical support, the assistance of AI is needed in addressing numerous challenges, such as how to quickly locate and deal with system fault (if any), how to allocate numerous resources according to the needs of industry users, what types and levels of resources shall be allocated appropriately, how to charge the allocated resources appropriately, what kind of application capability shall be allocated to users reasonably, and what kind of goods shall be allocated according to market changes.

## 2.2 Conclusion of BSS Evolution

By reviewing the main services of operators in the process of intergenerational upgrading of communication technologies and analyzing the evolution of their corresponding BSS, it can be seen that:

- Service innovation is the main driving force of BSS evolution and development. Communication services are initially determined by the capabilities of communication technology, mainly focusing on the capacity of suppliers, and then gradually transform to the needs of customers, combining with the overall business trend of the information industry (such as internetization).

- From the perspective of cooperation paradigm, BSS supports "internal and external double closed loops". By internally supporting the internal management and operation of enterprises, it can help enterprises better realize resource integration and optimization and improve internal collaboration and cooperation efficiency; by externally supporting the cooperation and development between enterprises and partners, it can empower the cooperation and development between enterprises and partners and improve the quality and efficiency of external cooperation.

- The evolution of BSS architecture is highly related to that of information technology in the same era, while information technology and communication technology mutually support and promote each other throughout their evolution process. For example, the internetization of Web1.0 has driven the development of broadband technology, the mobilization of Web2.0 mutually promoted by 3G and 4G technologies, and technologies such as IT cloudification and containerization have deeply affected the 5G communication technology.

# III. 6G Business Support Model

Considering the evolutionary impact of communication technology for operators' business model, the business models of communication operators at different stages are also studied and summarized in references [15,16,17,18,19]. As shown in *Figure 3-1*, in the 4G era and the previous intergenerational communication era, operators mainly focused on providing standard communication services to consumers and users at lower cost and higher RoI (Return of Invesetment). At this phase, operators, as value producers, provide professional connection services in a vertical communication field based on engineering functional platforms such as standard network device and OSS. With the simultaneous evolution of IT and CT, and the mutual penetration and influence, operators have begun to consider building digital infrastructure for enterprise users in the 5G era. The introduction and landing of the MEC architecture system and private network are the best evidence of this process. With the landing of communication technologies such as NFV, SBA and slicing, the communication capability gradually overflows outside the domain, which brings new value growth space for operators. As a general purpose technology, 6G strengthens the vision of value spillover, and operators need to build an ecological platform to promote the self-creation and self-evolution of the overall value of the society. Therefore, the development of 6G will promote operators to further integrate CT, IT, DT and OT, build a corresponding open ecology, and create a systematic value-added business model of total social value.

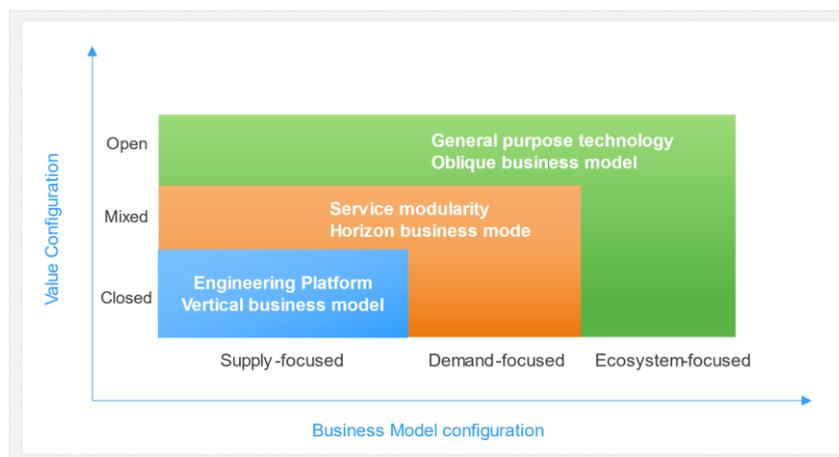

**Figure 3-1 Business Model of Communication Operators**

# 3.1 New BSS Requirements of 6G Typical Scenarios

In the past ten years, the mobile communication technology has developed rapidly, and the network performance indicators have achieved many breakthroughs. The main body of network connection has also shifted from personal communication to the Internet of Everything. Rooted in 5G and further improved in 6G, communication technology continuously enhances connectivity and certainty of network and promotes transition from mobile Internet to the Intelligence of Everything. This facilitates intelligent interconnection of people, the physical world and the digital world and provides more space for the diversity of 6G services. Although there are still many controversies about the architecture, key technologies and business processes of 6G, with the joint efforts of insiders, the industry has added scenarios such as  joint communications and sensing and proliferation of intelligence to the three typical scenarios of the 5G, and sorted out the typical services of 6G for 2030 [20,21,22,23].

- **Immersive cloud XR based on 6G ultra wireless broadband**

  Tbps peak rate, 10-100Gbps experienced rate, sub-millisecond latency, a tenfold increase in the density of 5G connections, centimeter-level localization, millimeter-level imaging, and E2E system reliability based on controllable error distribution can meet the communication requirements of various immersive cloud XR services. Immersive business with people-centric approach enables individuals to engage in work and daily activities through virtual environment and virtual characters anytime and anywhere; the XR expanded to industries such as smart security, smart cities, smart factories and data centers will also accelerate the comprehensive digital transformation of vertical industries and greatly enhance the productivity.

- **Universal IoT and Precise  Machine control based on 6G ultra-large-scale connection**

  Machine control is a kind of enhanced machine communication. With the arrival of 6G, combined with the integrated communication method of space and space, the ubiquitous IoT access and mutual communication of large-scale sensing devices and robots will be achieved. For example,Many industrial robots and autonomous robots have a strong demand for environmental awareness and human-robot communication. In the future, the cooperative machine control system will realize multi-dimensional cooperation between human, machine and environment based on more precise working environment and machine motion perception, more intelligent fine control and execution of more accurate control instruction. Its application scenarios mainly include smart factories, smart agriculture, smart cities, smart transport, smart energy and other fields. The business will be supported by tenfold increase in the density of 5G connections, centimeter-level localization, millimeter-level imaging, and other 6G network performance advancements.

- **Holographic communication based on 6G ultra wireless broadband**

  Holographic communication service is an overall application solution of data collection, coding, transmission, rendering and display of high-immersion and multi-dimensional interactive application scenario based on naked-eye holographic technology. Consisting of the whole E2E process from data collection to multi-dimensional sensory data restoration, it is a highly immersive service form with high degree of natural interaction. With the comprehensive improvement of 6G network communication capability and the support of high-resolution terminal display devices, holographic communication services will be greatly developed and widely used in many fields such as culture and entertainment, medical health, education and social production, providing multi-dimensional interactive experience, holographic digital management, high-quality portrait interaction and other application services.

- **Ultra-high precision positioning based on 6G joint communications and sensing**

  The 6G network has sensing function that allows it to provide active positioning service for communication objects and passive positioning service for non-communication objects. It uses communication signals to realize sensing functions such as detection, positioning, recognition and imaging of the targets, obtain the information of the surrounding environment, and help to complete the digital virtualization of entities in the environment. Typical scenario services include precise positioning, environmental reconstruction, security imaging, drone delivery, and autonomous driving.

- **Enhancement of interpersonal communication by brain-computer interface based on 6G proliferation of intelligence**

Relying on the future 6G mobile communication network, it is expected to make breakthroughs in such new research directions as emotional interaction and brain-computer interaction (brain-computer interface). Intelligent systems with perceptual, cognitive and even thinking capabilities will completely replace traditional interactive intelligent devices, and the hierarchical relationship between people and agents will become human-like interaction that involves emotions, warmth, and equality. The intelligent system with emotional interaction capability can monitor the psychological and emotional state of users through voice dialogue or facial expression recognition, and adjust the emotion of users in time to avoid health hazards; by manipulating the machine through the mind or brain to replace some functions of the human body with the machine, it can also make up for the physical defects of the disabled, maintain an efficient working state, efficiently acquire knowledge and skills within a short time, or realize "lossless" brain information transmission.

From the above typical 6G services, it can be seen that by building a 6G network with native intelligence, operators can upgrade the network connectivity to information perception and communication capabilities, thus supporting digital business innovation in thousands of industries more efficiently. Because of the more open 6G network service capability, operators need to consider some common characteristics presented by 6G services:

➢ As the basic underlying demand, the upgrading of network capability will ensure that complex and diverse network business scenarios can be built on standardized communication technologies.

➢ The ultimate network experience and proliferation of intelligence brought by 6G enable network services to focus more on higher-level innovations such as modes and scenarios; as a result, pure communication operators are no longer the core of ecological cooperation.

➢ The breakthrough of the bottleneck in network connection has made 6G a solid infrastructure base, the concept of "network as code" has been implemented, and ecological cooperation and collaborative innovation have become important themes in the 6G era. It can be seen that the 6G business needs an open 6G network to provide the communication capability of intelligence of everything, and it also needs to combine domain knowledge and integrate emerging technologies to empower digital intelligence upgrades for thousands of industries.

Faced with these changes, operators must further improve digital efficiency and empowerment, and transform from communication service provider (CSP) to digital service provider (DSP) before the advent of the 6G era. The improvement of digital efficiency means that operators need to optimize the process and framework of BSS, introduce new technologies, continuously improve the efficiency and scalability of core businesses, and complete the transformation from omni-channel customer center to automated network operation center, including intelligent demand decomposition, ultra-automated service experience, all-round service efficiency evaluation, adaptive service optimization and other business support closed loops. Due to the improvement of digital empowerment, operators have to explore how to support a new digital ecosystem in BSS, extending from the communication field to manufacturing, automotive industry, health industry and smart cities. Operators will go beyond "pipeline business" and establish "platform business model", thus completing the transformation from "building competitive advantage based on network resources" to "empowering ecology based on network resources". To this end, operators need to build a more refined product and service system in conjunction with the ecology to promote business prosperity; BSS needs to support the rapid iteration and convergence of business cooperation mode to meet the demands for rapid introduction by partners and opening-up.

## 3.2 Reference Framework of BSS

### 3.2.1 Business Support Framework from Telecommunication Organizations

With the rapid development of the communication industry, BSS of operators is constantly incorporating technical standards of IT industry, establishing its own ecological organization, and improving the relevant technical standards in different periods with business development.

Around 2009, in order to meet the industry demands amid the ICT convergence trend and the development of digital media and digital services, TM Forum revised the eTOM model, removed the label of telecommunications and established the BFM (Business Process Framework) model which eventually became the business architecture in ODA (Open Digital Architecture) [24].

Although the IT support system of operators in the early stage has developed rapidly with the business growth, many problems have been exposed in the process of construction, deployment and application and some have seriously affected the sustainable development of the system. The construction and operation of the operations management system have a direct impact on the overall cost, management and service of the communication operations. Therefore, in the process of participating in enterprise information construction, operators have also begun to apply the general enterprise IT service management methods and standards. Among them, the ITIL, a process-centered IT management industry standard, is a typical standard and method. IT service management oversees the full life cycle of IT services, which involves human resources, organizational structure, management, process and technology, and other aspects, including pre-research, planning and construction, operation & maintenance of IT [25].

On February 27, 2023, the industry-wide initiative plan of GSMA Open Gateway was grandly released at the Mobile World Congress (MWC 2023). GSMA Open Gateway is a universal network API framework, which aims to provide a universal access interface, facilitate developers and cloud service providers to access operators' networks faster, and enhance and deploy related services [26]. Currently, in the plan of GSMA Open Gateway, eight universal network APIs have been launched, including SIM card exchange, QoD, device state (access or roaming status), code verification, edge site selection and routing, code verification (SMS 2FA), operator billing-withdrawal and device location (verification location). It is expected that more APIs will be launched in 2023 according to the initiative plan.

With the construction of 5G network, the communication industry is becoming the vanguard of the digital era, and its information system has also changed from a business supporter to a business enabler, becoming a tool for operators to fully empower the digital society. Therefore, facing the evolution of 6G network, operators need to define their business support capabilities more comprehensively and systematically.

On the whole, TM Forum defined the open digital architecture (ODA)[15] in 2019, and provided a possible system architecture reference for 6G BSS. This architecture intended to replace the traditional operation support system (OSS) and business support system (BSS) with a new method, and provides the future-oriented blueprint, language and a set of key design principles accepted by the industry. ODA mainly includes five parts:

- **Business Architecture:** Including key business process multi-layer model (eTOM) that supports efficient and agile operation, and capable of mapping business capability and value stream.

- **Information System Architecture:** Including functional architecture and data architecture (SID) that support loose coupling, being capable of providing standard information definitions among operators, suppliers and other partners.

- **Implementation Architecture:** Including more than 50 REST-based Open APIs, and capable of realizing the interoperability of standardized IT system; supporting reused and simply integrated ODA components. Standardized data models can also help achieve AI operation and maintenance.

- **Deployment & Runtime Environment:** Using Canvas to support the running of plug-and-play ODA components, providing standard technical framework and DevOps support, and making verification by continuous deployment in laboratory test environment.

- **E2E Governance:** Providing related principles, design guidelines, metadata, and agile management tools covering the full life cycle of the architecture.

Through the collaboration of these five parts, operators can define their business architecture according to the ever-changing business model, and create a digital BSS capable of efficient, agile and automatic operation & maintenance like building blocks.

Combined with the domain-based organization form of operators, the "customer-centric" operation mode is abstracted for the future, and the core part of ODA, i.e. eTOM business architecture (as shown in *Figure 3-2*) is constructed, which emphasizes the domain-based business process and divides the overall process into two parts from the perspective of enterprise management: strategic planning and operation.

| | Strategy and agreement | Infrastructure life cycle management | Product life cycle management | Operation support and preparation | Business opening | Business guarantee | Billing and revenue management |
|---|---|---|---|---|---|---|---|
| Market / Marketing | | | | | | | |
| Customer | | | | | | | |
| Product | | | | | | | |
| Service | | | | | | | |
| Resource | | | | | | | |
| Partner | | | | | | | |

| Enterprise management | Strategy and enterprise planning | Finance and asset management | Knowledge and research management | Stakeholder and external relationship management | Enterprise risk management | Enterprise process management | Human resource management | General process management |
|---|---|---|---|---|---|---|---|---|

**Figure 3-2 eTOM Model of TM Forum**

Although the information system architecture under the ODA system provides a reference for BSS implementation in the communication industry, the ODA only presents the logical framework of the key function grouping L0 of BSS. It also identifies and captures five decoupled functional modules related to the work done by the organization, and connects and integrates them through the Open API (as shown in *Figure 3-3*). For the L0-L1 expansion, the business process in light of the eTom business model needs to be further refined.

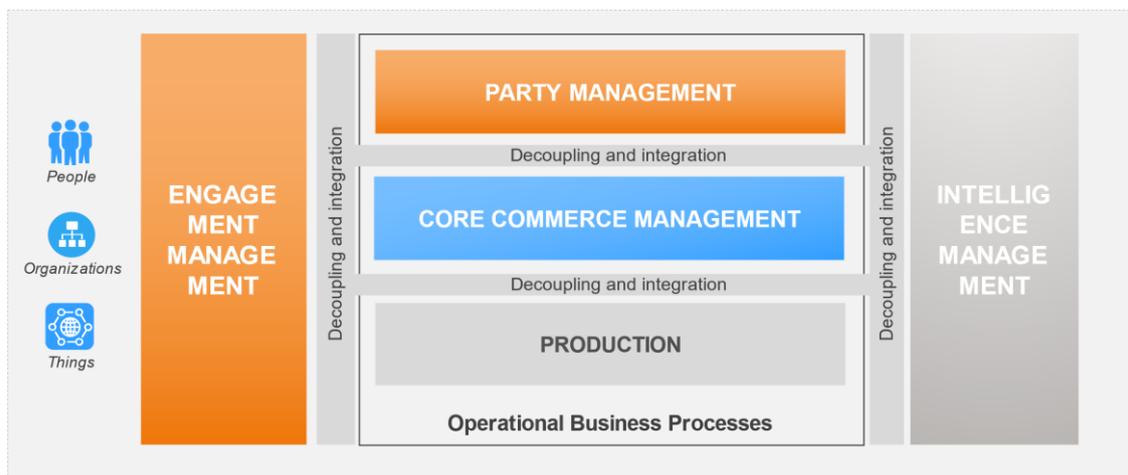

**Figure 3-3 ODA Functional Architecture of TM Forum**

In the future, the 6G BSS system can further evolve on the ODA functional architecture, and introduce corresponding capabilities in combination with the development of the Internet to build a more open and flexible digital business support capability from the bottom up, thus establishing an operator-centered ecosystem to accelerate the development of the digital economy. This system provides the following prospects for the business development of communication operators in the 6G era:

- Build a digital infrastructure featuring computing and network convergence based on network resources, thus developing hierarchical services and products from the bottom up and enable operators to expand their digital capabilities of external empowerment.

- Remodel the whole process in a customer-centric approach, and attach importance to customer service experience and efficiency evaluation.

- Process marketing capability support, supporting the landing of diversified business models.

- Promote the collaborative communication of B domain, O domain and M domain, and build an efficient, agile and open digital BSS.

## 3.2.2 Business Support Reference from Internet Industry

Metcalfe, the father of computer network, once pointed out, "The value of a network is the square of the number of nodes in the network and is proportional to the square of the number of connected users of the system". The Internet demonstrates the power-increasing relationship between network scale and network value, and based on the connection of people, greatly magnifies the total ecological value through open cooperation. Reference [17] focuses on typical Internet business in the Web 2.0 era, and gives 4C business model classification method (Connection, Content, Context, Commercial) from the top, representing social network, content service, information retrieval and e-commerce business respectively. In the era of Web2.0, focusing on the core service value, enterprises gather data via platforms, and try to realize traffic monetization through advertisements, contents, commissions and other ways. In the process of supporting consumers' business, the support systems of Internet companies constantly polish their technical capabilities, forming a cloud-native IT system and DT system. In addition, they also build a middle platform business support system for developers, which not only effectively supports their own SaaS development, but also greatly empowers the corresponding cloud computing business based on IaaS and PaaS.

As a result of Web2.0's centralized nature, the data from all aspects is collected by a few dominant players due to mergers and restructuring within the internet industry. This brings huge profits for the industry giants, but meanwhile has created social injustice and insecurity. Then, Web3.0, based on blockchain, DAPP, semantic web and other technologies came into being, with the prospect of breaking the monopoly mechanism built on the platform in the Web2.0 era, and a large number of bottom-up innovations are quietly taking place. In this process, the research on business model is also in full swing. At present, the industry generally adopts bottom-up induction [27,28,29,30] to list various business models that have been implemented (such as charging according to License and blockchain service). On this basis, reference [31] summarizes the business model as shown in *Figure 3-4*. Its core idea is to promote multi-party cooperation and jointly create ecological prosperity around the core value model of key roles, based on consistent blockchain model, ecological expansion model and incentive distribution rules. In essence, Web3.0 is to build a trusted digital society by imposing a wide range of constraint rules on a low-trust network through technology. From the perspective of value creation and value flow, Web3.0 inherits and continues the successful paradigm in Web2.0. With the implementation of production modes such as DAO, more innovations can be expected in the future. The related digital capabilities supporting Web3.0 businesses are divided into technologies of the basic layer, platform layer and application layer. The basic layer technology consists of the convergence technology of blockchain, including distributed ledger, consensus algorithm, cryptography technology, smart contract, distributed storage, cross-chain, etc. The platform layer technology includes AI, big data, extended reality (XR), cloud computing, rendering, 3D modeling and so on. Application layer technologies include a wider variety and offer more possibilities, suggesting huge potential for the integration and development of various technologies. Therefore, practitioners of Web3.0 are also providing related services and improving the corresponding BSS from the bottom up around decentralized enabling technology.

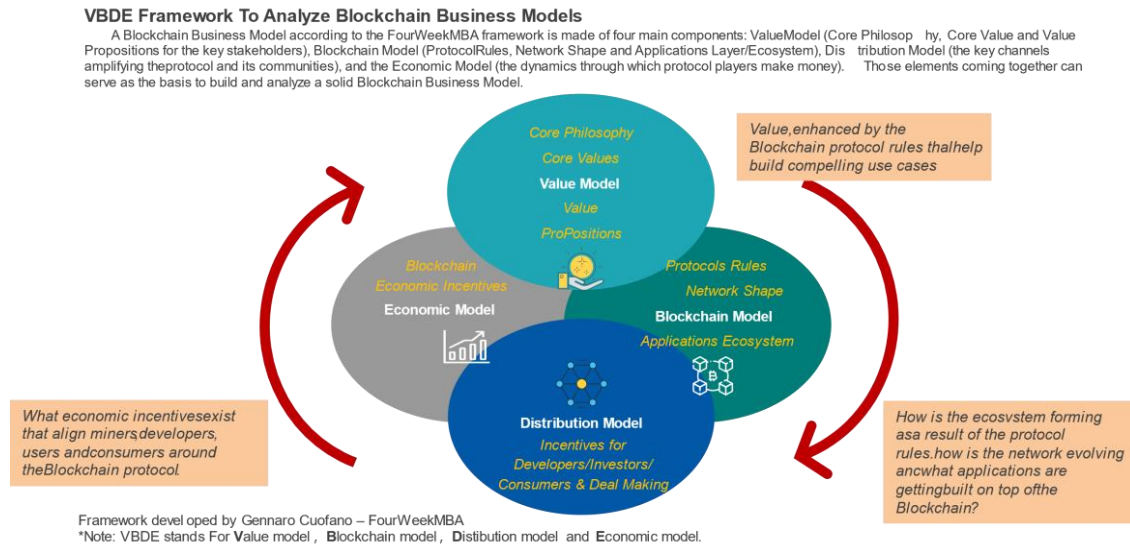

**Figure 3-4 Business Model for Web3.0**

In terms of connection enablement, the communication technology has built a more stereoscopic and prosperous cyberspace than the Internet. How to maximize the ecological value of operators based on the powerful cyberspace? 6G Flagship and other organizations [16,18,19] once introduced the 4C business model framework [17], which was applied to Web2.0, into the communication field (as shown in *Figure 3-5*). The purpose of this was to better cope with the increasingly diversified products and services of communication operators. From the business development of the Internet industry, the following points are worthy of reference for the communication industry:

● Consensus on core values regarding to its own superior resources.

● Keep an open mind in cooperation to release more business potential.

● Change the business support from internal service to customer service, and focus on customer experience.

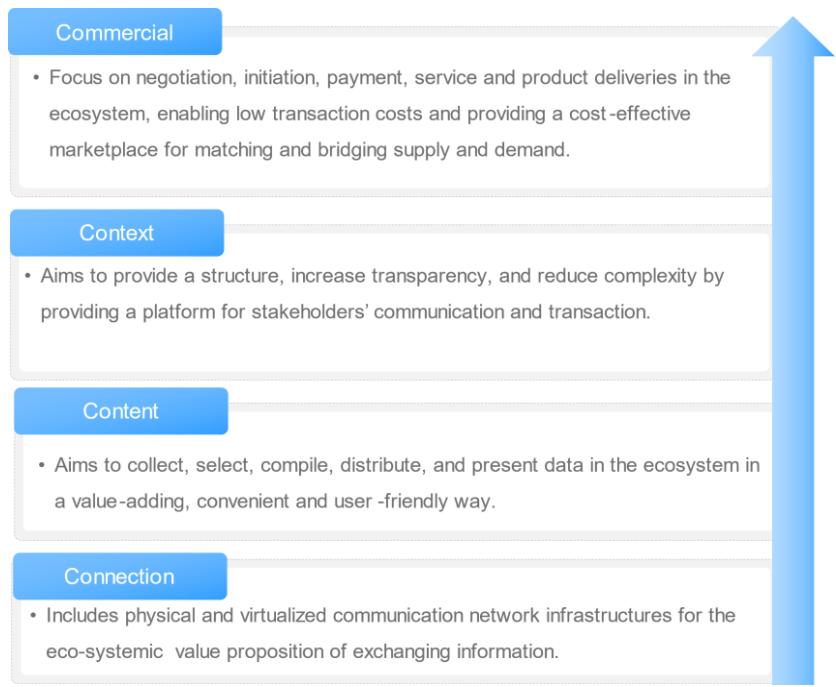

**Figure 3-5 4C Model of Communication Business**

In practice, operators have also expanded their flow management business with the help of the mobile Internet Connection capability provided by 4G, and began to gradually introduce the cloud native and big data of the Internet to improve the agility and intelligence of consumer business support, and then further expanded the industrial-level digital business through the cloudified middle platform service under the capability support of the 5G Internet of Everything. For the future 6G era, operators should not only consider the continuous influence of Web2.0, but pay attention to the new business model and business innovation brought by Web3.0. Their digital BSS shall be enhanced in the following aspects:

- With the innovation of the underlying technology from 4G, 5G to 6G, the standard connection value of communication operators is improving, including horizontal (perception capability, computing power integration capability, ICT convergence, etc.) and vertical (extreme improvement of network communication performance indicators) value expansion. As a result, the underlying core value construction is no longer limited to the network, but extends to the dimensions of computing power, perception terminal and storage power.

- As a general purpose technology, connection has increasingly prominent value in enablement. Operators can use this core competence to integrate more technologies besides data, including intelligent collaboration, service connection, tool integration, etc., and build a developer ecosystem to enhance their business innovation capabilities.

- Under the decentralized business model, it is particularly important to establish a digital ecosystem with operators as the core. Establishing a digital support system with incentive mechanism and efficient operation for business innovation will drive operators to continuously provide high-level value-added services and stimulate business innovation.

### 3.2.3 BSS Evolution and Planning of Operators

With the advent of the digital economy era, the communication industry has been making proactive efforts since the 4G era to cope with the rapid changes in the market and technology. In the current 5G era, with the saturation of the personal communication market, the enterprise market has indisputably become the growth pole of operators in the future. However, enterprise services such as SD-WAN, enterprise security and network slice need to be built on the combination of virtualization and various network functions (RAN, core, transmission and cloud), and are not provided by isolated network domains. In addition, the needs of different enterprise customers and the ways of business cooperation are quite different. Therefore, digital service providers must optimize their operations to ensure the improvement of delivery cost and delivery efficiency.

To respond to these changes, the global mainstream BSS manufacturers and even operators have put forward a series of corresponding technologies, products and solution systems. For example, the established communication BSS manufacturers Amdocs, Ericsson, Huawei, AsiaInfo, the international operator Verizon, and the three major domestic operators in China have provided corresponding solutions for the common needs of global operators.

The digital brand experience support system provided by Amdocs is a digital customer management, commercial and monetization solution based on the existing BSS, which is specially designed for the digital brand needs of service providers, who need to provide digital experience for their customers while maintaining agility, innovation and rapid launching capability. This solution helps operators to streamline the closed loop of their business support by pre-building business and technical processes across the full life cycle, including customer care, business, ordering and monetization. In addition,

Amdocs provides cloud migration service on BSS, which is planned and executed by Amdocs in cooperation with AWS, involving a large number of BSS data, for the purpose of providing better personalized user experience, improving self-service adoption rate and greater flexibility.

In response to the transformation of communication service providers to digital service providers, Ericsson has built a new customer-centric BSS for operators. By redefining the multiple layers of digital customer enabler, digital business enabler, service management and network & cloud infrastructure from top down, a multi-level value monetization system is established for operators to unleash business vitality.

AsiaInfo has proposed the concept of "global virtualization, global intelligentization and global perception" ("three global domains" for short) of 5G to enhance the technical evolution of 5G private network, 5G network intelligentization, computing network and customer experience management (CEM) towards 3GPP 5G Advanced. They focus on three major technology modules (cloud network, IT and digital intelligence) and the technical evolution of Top X core technology of three major middle platform systems (technical middle platform, data middle platform and AI middle platform), as well as the continuous evolution of BO convergence technology.

In response to the global wave of digital economy, international operators have developed a variety of technologies and solutions for enterprise support systems, such as digital transformation solutions, cloud solutions, mobile solutions, and IoT solutions.

Beyond the enterprise market, the metaverse business based on Web3.0 may reshape all aspects of personal life. The first obvious business opportunity brought about by the metaverse for communication operators is the upgrade of connectivity. With the support of the 6G network, the metaverse business will move from those such as simulation activities, customer seminars, webinars, and digital twins for corporate operation to higher-level metaverse application scenarios such as immersive XR, holographic imaging, and sensory interconnection. In the face of the changes brought about by the metaverse, some operators do not hope to be simple connection providers. For example, Deutsche Telekom recently cooperated with Telecom to bring the Korean operator's ifland Metaverse platform to Europe, allowing users to create virtual images and establish virtual conferences in the "New World". For operators, the metaverse will be an opportunity to establish partnerships and do more than connection service providers. Operators may have the opportunity to provide a "business platform" to use the combination of product catalog, bills, charging, partner management, care, sales and marketing to power the innovative businesses based on the metaverse. Metaverse has opened up a variety of new methods for the business innovation of operators. In the metaverse based on Web3.0, the potential for new business model, personalized contextual advertising & services, and value-added sales of digital content are enormous, but this also highlights the need for the existing BSS to provide a high degree of openness, adaptability and flexibility.

In order to better support the development of metaverse business under Web3.0, a TM Forum catalyst project jointly completed by multinational operators proposed to introduce a customer digital avatar driven by AI with decentralized digital identity and data ownership into the existing BSS, in order to provide new digital contact points for customer care and service agents, thereby comprehensively improving customer experience, employee experience, use experience and multiple experiences, and then strengthening the customer-centric sustainable business development model. Chinese operators have also defined a ubiquitous computing-aware network resource layer, a game-based computing power engine layer, an immersive social experience layer, and a hybrid reality business layer in the metaverse BSS from the bottom up. It can be seen from this plan that operators have borrowed ideas from the Internet companies to support the co-construction of infrastructure, the creation of development ecology, and the upgrade of user experience in a more inclusive manner. From operators' current business exploration of the metaverse, it is observed that communication operators are accelerating their transformation into digital service providers, so creating a new BSS for the 6G digital innovation has also become part of their agenda.

## 3.3 6G Domain-based Business Support Model for Digital Innovation

From the open digital architecture developed by the industry standard organization TM Forum, to the theoretical research on communication service models under Web2.0 in the industry, as well as the new topics brought by Web3.0 and 6G for operator business, and even the transformation practices of operators themselves to digital service providers, a possible evolutionary path for BSS capabilities in the 6G context can be summarized.

First, the 6G network will greatly enhance the connection advantages of operators in infrastructure resources, enabling operators to adjust, integrate and enhance computing power with the support of the network, build blockchain based network infrastructure for Web3.0, and widely connect infrastructure providers through the 6G network so as to integrate network-wide resources. Second, operators can open up the CT capabilities of 6G network like Internet companies, introduce IT, DT, and OT capabilities, and create assembled full stack digital technology services for production developers, in order to support the innovative development of digital business. Third, based on ubiquitous computing-aware network and full stack digital technology capabilities, operators can enhance their own digital business innovation, and also work with partners to promote industrial digital business innovation. Finally, operators need to establish corresponding ecological cooperation at different levels such as resource supply, production & development and business innovation based on their core capabilities, in order to ensure the sustainable development of the 6G business. A customer-centric total experience and integrated operation and empowerment will run through the above domains, and the corresponding 6G BSS will further improve the operation efficiency of operators in response to the above needs and upgrade the business support ability for external empowerment.

To improve the digital operation efficiency of operators themselves, 6G BSS needs to be continuously upgraded in the following two aspects:

- **Transition from internal-user-supported to customer-centric:** Technologies such as intention recognition, new SLA evaluation system, data fabric, digital twin and ultra-automation should be introduced to build unified digital identities around customers and optimize customers' service experience.

- **Whole-process collaboration, enabling integrated and automated operation:** New technologies such as digital intelligence, cloud native and Devops for operator-oriented business operations and support systems should be introduced to continuously improve the operation efficiency and scalability of core business.

To improve the digital empowerment efficiency for operators, 6G BSS needs to fully support ecological cooperation in a more open form:

- **Accurate positioning of foundation core values:** In the intergenerational replacement of communication capabilities, the carrier of communication core values has completed continuous upgrades from voice, text, traffic, video, and future holographic communication. With the introduction of technologies such as computing and network convergence and Web3.0 into 6G, future-oriented computing-aware network resources and decentralized Web3.0 support capabilities have also become new value carriers.

- **Technology convergence drives product innovation:** Due to the convergence and development of IT and CT technologies, operators can not only provide their powerful communication technology services, but also introduce technologies such as cloud computing, big data, AI and industrial interconnection operations to offer developers digital service production capabilities, with the services required by decentralized applications taken into account to guarantee user data ownership under Web3.0.

- **Value creation drives business innovation:** The general purpose technologies and underlying resources provided based on the digital production layer can effectively promote the operators and ecological partners to complete business innovation. In addition to trusted transactions of resources such as computing network and data, developers may obtain digital tools, end users can get corresponding digital products, and users are empowered to realize value through their own digital assets in the era of Web3.0.

- **Connecting participants to promote ecological cooperation:** All business innovations shall be based on effective business contracts of all participants. The ability to support innovation of different business modes and trusted transaction will be the key to the future ecological construction of operators. In particular, those of interest should pay attention to the effective utilization of the decentralized autonomous organizations emerging under Web3.0 to establish the ecosystem advocated by operators.

In order to implement the above design elements and adhere to the principles of decoupling and collaboration, this White Paper proposes a domain-based business focus and collaborative cooperation model for digital innovation (see Table 3-1 below) to support the digital innovation of 6G business.

**Table 3-1 Operator Business Model for the 6G Era**

| Digital innovation | Design points of BSS | Domain-based decoupling |
|---|---|---|
| Improvement of digital operation efficiency | Transition from internal-user- supported to customer-centric | Experience |
| | Whole-process collaboration, enabling integration and automated operation | Operation |
| Digital empowerment efficiency | Accurate positioning of basic core values | Resource |
| | Technology convergence drives product innovation | Production |
| | Value mining drives business innovation | Business |
| | Connecting participants to promote ecological cooperation | Ecosystem |

- **Experience domain:** This domain will focus on customers, identify breakpoints and optimizable points in business processes, and improve customer experience, employee experience, user experience, and multiple experiences. It will focus on total experience optimization of service processes.

- **Operation domain:** This domain will break down the institutional and technical barriers of various domains of operators, and achieve comprehensive automation of operation processes for safety production requirements and business use processes. It will focus on automation and intelligentization of operations across all domains.

- **Resource domain:** This domain will determine the core basic resources and corresponding resource management methods of operators, with a special focus on the overall resource utilization rate, as well as lean resource integration and operational management of resources.

- **Production domain:** This domain will integrate various digital production capabilities and produce corresponding digital tools or processing lines. It will focus on the production and R&D of digital products, and provide ecological support for creators.

- **Business domain:** This domain will determine the business process and operation mode, formulate corresponding product plans, and provide core functions to ensure the best user experience. It will focus on deliverable services and products, and ensure the full life cycle management of products and commodities.

- **Ecosystem domain:** This domain will build a communication channel for ecological partners with operators as the center and provide real-time incentives and scenario empowerment through capital operation and other methods to quickly acquire the resources and capabilities required for cooperation. It will focus on ecological business model design, revenue generation, and formulation of commercial rule.

# IV. Overall Vision of 6G BSS

This White Paper holds that 6G, as a general purpose technology, will strengthen the vision of value spillovers, and operators also hope to become the core of the digital service ecosystem with the help of 6G capabilities, so as to secure an advantageous position in the Internet digital upgrading process. So, 6G BSS shall be able to construct an ecological system in a fully open form, and promote the self-creation and self-evolution of the overall digital economic value of the society. The accelerating transformation of digital business and the upgrading of capabilities driven by 6G technology from the bottom to top require operators to focus on different layers and collaborate on digital innovation. This 6G domain-based business support model will continuously strengthen and improve the digital operational efficiency and digital empowerment efficiency of operators. It also enables the development of BSS with DOICT as its technological basis, thereby realizing a series of visions, such as ecological valuation, business diversification, production flexibility, resource sharing, scenario-based experience, operation automation and native safety and security, and enabling the 6G digital business innovation. These are shown in *Figure 4-1*:

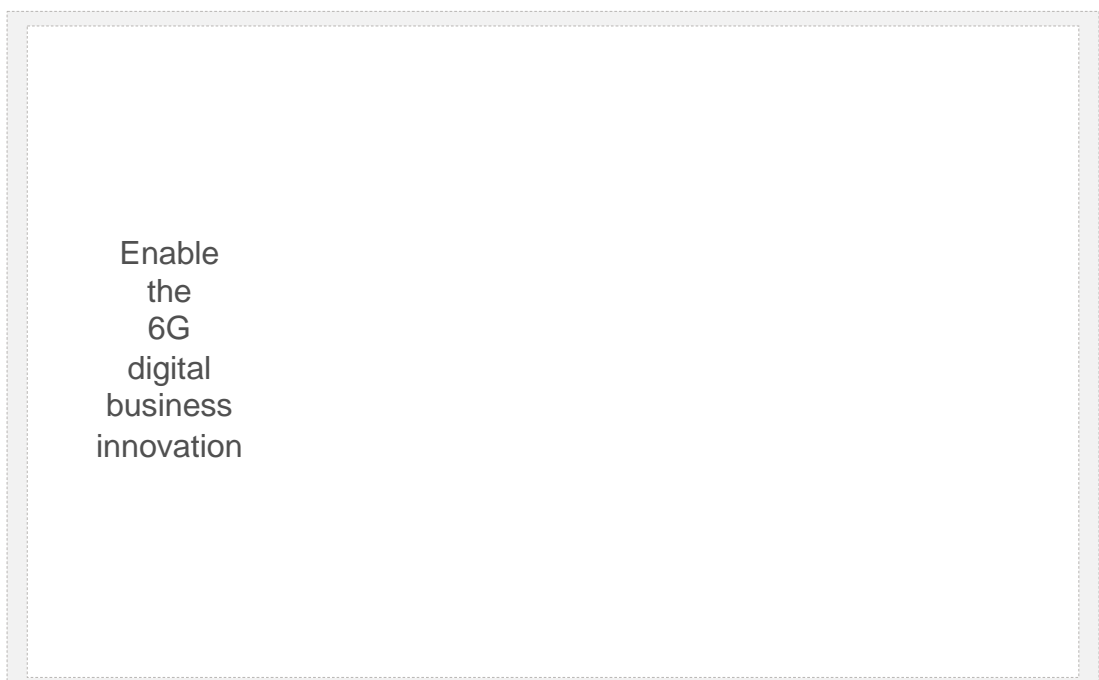

**Figure 4-1 BSS Vision for 6G**

- **Scenario-based experience:** In the experience domain, diverse experiences are provided based on different business scenarios to meet the needs of total experience of intelligent interaction and virtual-real interaction required by different role objects during enterprise business activities, which are mainly reflected in:

➢ The objects satisfied by the digital experience include employees, corporate customers, partners, suppliers, consumers, etc., focusing on the system platform and customers' zero-touch collaboration and business intention perception capabilities.

➢ In terms of digital experience interaction mode, it has expanded from traditional human-computer interaction, PC terminal and mobile terminal to total experience modes such as brain-computer interaction, virtual-real interaction between intelligent agent and twin aggregate, metaverse and super APP.

➢ Support customer digital avatars, digital virtual humans, and achieve an omni-channel, full-process business intent driven experience. Being able to convert experience orders such as "I want machine-checked connections" into seamless on-demand services for product management provided and charged by a digital innovation platform.

● **Operation automation:** In the operation domain, focusing on the automation, process mining, and security of BSS business processes can be expanded to external enabling meeting the future business ecology monetization, to achieve the integration of B2B2X business ecology. Through the E2E business process automation, operations of resource, production, business and ecology are connected in series to improve digital operation and empowerment efficiencies, which is supported by security technologies such as platform engineering, blockchain and zero trust, embedded in the E2E business processes of enterprise production, and coordinated with intelligent process mining and optimization to ensure the credibility of the entire process.

● **Resource sharing:** In the resource domain, network resources (connection, perception, positioning, etc.), computing power resources (CPU, GPU, etc.), storage resources (memory), and related IT high-level computing power resources can realize sharing, reuse, elastic and efficient resource management

[32,33] through cloud-edge-terminal distributed and container technologies, which are mainly reflected in:

➢ The integration of physical resources at different regions and levels.

➢ The integration of cloud-edge-terminal three-dimensional ubiquitous logical resources.

➢ The integration of heterogeneous resources brought by different computing hardware and diverse chip architectures.

● **Production flexibility:** In the production domain, new technologies such as data fabric and digital twins are introduced as needed with the support of a cloud native component-based framework. The technical architecture is easy to expand, technical elements are easy to integrate, and technical capabilities are easy to invoke. Zero code/no code and visual drag-and-drop tools are provided for developers to simplify the application complexity of these technologies [32]. Through production domain integration, various types of digital production capabilities are integrated and corresponding digital tools or processing lines are produced, focusing on the integration of digital product production and R&D, and providing ecological support for creators to enable rapid transformation and collaborative production of multiple technologies, processes, forms, and units.

● **Business diversification:** In the business domain, integrated communications and sensing, air-space-ground integration, proliferation of intelligence, and high-precision perception and positioning further diversify the business and scenarios. In terms of diversified capacity building, the key points are:

➢ Extracting common features of business capabilities to form a digital business virtual network element and service unit orchestrator [34]; integrating and solidifying through business service descriptor (BSD), and building standard business processes to achieve visualization, agile development and innovation, flexible deployment and scenario operations of business.

➢ Building a new business service evaluation system that integrates QoE (Quality of Experience) and QoS (Quality of Service), which can be used as a model definition for new communication services to analyze and measure visual/tactile sensing accuracy, experience, etc.

● **Ecological valuation:** In the ecosystem domain, B2B2X represents a significant breakthrough in the past model for operators. The "chain connection" model based on simple upstream and downstream cooperation of "service provision, service procurement", has evolved into a "network interconnection" model of multiple types of cooperation and collaboration [32]. As operators of value network, operators have also shifted from "solution promoters" to "solution creators", releasing the new paradigm of ecological cooperation and systematic capabilities such as customer security and privacy, data flow, trusted transactions, decentralized autonomous organizations. BSS shall be strengthened to build a benign ecosystem, and provide protection for the sustainable and healthy development of digital systems.

# V. 6G BSS Potential Technologies

Generally, BSS presents an epochal business development trend in the 6G era, which requires the support of a series of future-oriented IT technologies. Based on the advancement timeline of 6G and the evolution characteristics of BSS, and for the reference of readers, some technologies that may serve as the future IT support for BSS have been identified and examined. These technologies, yet to be developed, are expected to be available around 2030. *Figure 5-1* summarizes the recommended potential key IT technologies in the planning of 6G BSS.

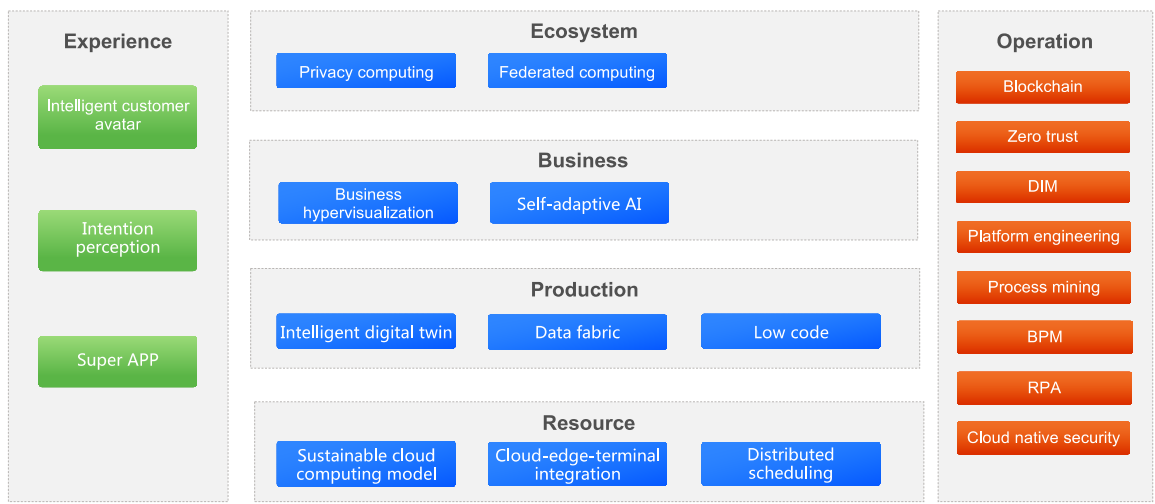

**Figure 5-1 Potential Key Technologies of 6G BSS**

6G BSS will be a system for application ecosystem and value chain, so the sharing of resources among applications, scenarios, and partner support systems is very important. Based on this requirement, a sustainable cloud computing model needs to be introduced in the resource domain to support a cloud computing environment that is not entirely owned and controlled by operators. In addition, due to the concept of 6G sensing as a network and its E2E capabilities for vertical industries, cloud-edge-terminal integration technology is naturally introduced to share cloud-edge-terminal computing resources. It is also necessary to introduce distributed scheduling technology to achieve unified resource scheduling.

Due to the complexity and importance of the system in the 6G era (considering its involvement in core production and personal health), it will be necessary for individuals to exert their personal initiative to achieve democratization of manufacturing. Based on such premise, intelligent digital twins and data fabric technology may be introduced into the production domain. Intelligent digital twin technology can provide comprehensive perception and prediction capabilities for system managers, thereby essentially improving the controllability of the system; furthermore, data fabric technology can provide data integration capabilities of low cost and high timeliness, which will be a good foundation for monetization of operational data.

At the level of business domain, considering the requirements of the Internet of Everything and interactive perception in the 6G era, supporting business diversification will become the development trend of BSS. Therefore, the ability of AI will become the key for operators to stand out from the competition. To achieve business diversification, it is necessary to introduce self-adaptive AI technology to enable operators to avoid long training and parameter adjustment processes when using AI technology. It is particularly crucial to allow AI technology to adapt to the business. Due to the complexity of business and the diversity of participants, it will be very important to introduce business hypervisualization technology to respond to the market and application ecology in time. Similarly, in order to reconcile the need for system participants to use data in fierce competition and the growing awareness of various stakeholders to data privacy, the introduction of privacy computing technology will be inevitable.

Experience in the 6G era requires a more natural human-computer interaction. Therefore, introducing an interactive mode based on intelligent customer avatars in the business domain will enable operators to establish a more humanized and efficient interactive connection with customers, thereby achieving the ideal of collaborative human-computer interaction. This connection also needs to overcome the inherent vagueness and ambiguity problems of natural interaction models, so intention perception technology needs to be introduced synchronously to truly capture the customer's intentions and provide timely and appropriate responses.

Due to the characteristics of 6G BSS such as system participants, application ecological environment, Internet of Everything and system scale, the automation and security of operations will be crucial. Therefore, the system needs to establish a stronger security and operation system, in which the zero trust security control and native safety and security based on blockchain trusted identity authentication shall be introduced to establish a security system, and the introduction of super automation, platform engineering and digital immune system shall be an issue worthy of serious consideration so as to improve the overall operation efficiency of the system.

# 5.1 6G BSS Technical Framework

The 6G BSS technical framework adopts a cloud, edge and terminal architecture, as shown in *Figure 5-2*. Each part is designed based on a layered architecture and can simplify complex issues. Based on the principle of single responsibility, each layer of codes performs their respective duties. Based on the design methodology of "high cohesion, low coupling", the interaction between relevant layers is achieved, thereby improving the maintainability and expandability of the platform.

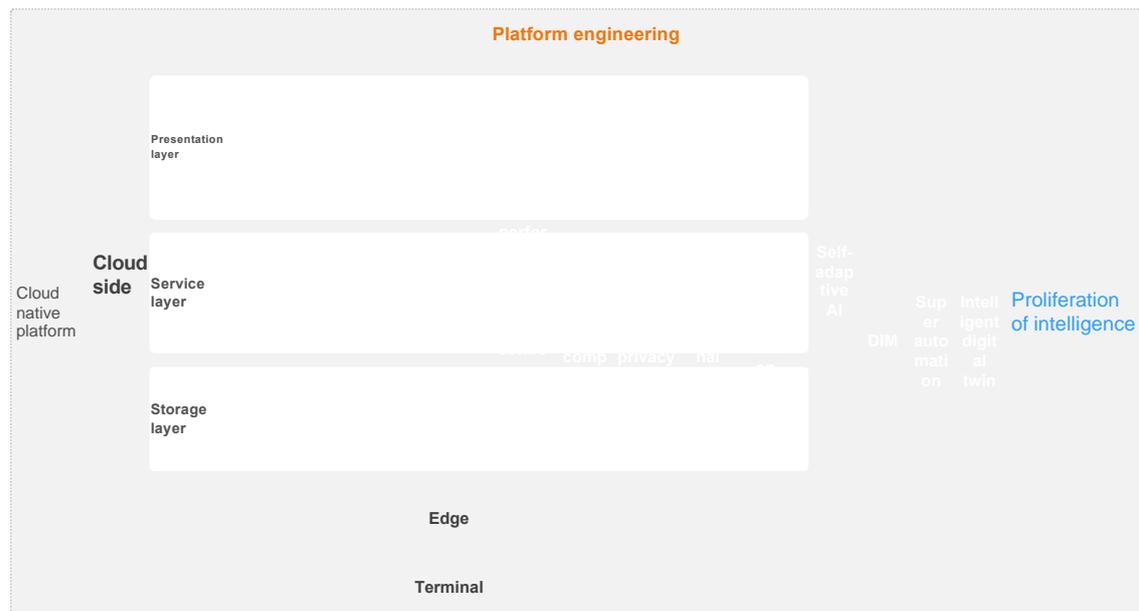

**Figure 5-2 6G BSS Technical Framework**

The entire technical framework follows the platform engineering philosophy in order to optimize the developer experience and accelerate the product team's speed for creating value for customers. For different business domains, BSS, as an industry cloud platform, can provide industry modular capabilities to support industry application scenarios by combining SaaS, platform as a service (PaaS), and infrastructure as a service (IaaS). The platform truly realized "continuous and ubiquitous learning and perpetual updates" across different business ecosystems, offering AI services and applications to every end user and allowing real-time and reliable AI intelligence to become a faithful partner for every individual, family and industry, thus achieving proliferation of intelligence.

When any interested party takes any type of action, BSS will generate "observable data" containing digital features, such as logs, traces, API calls, dwell times, downloads and file transfers and feed back all observable characteristic data in a highly coordinated and integrated manner to create a decision-making cycle, thereby improving the effectiveness of organizational decision-making and the application observability capabilities. The digital immune system will create an excellent user experience (UX) while reducing system faults. The platform using the digital immune system can reduce system downtime by up to 80%, and the reduced losses will directly translate into higher revenue. The cloud-edge-terminal integrated BSS provides unified application and operation management for business scenarios, and will become a fundamental platform for digital transformation in various industries, enabling in-depth empowerment for digital transformation and intelligent upgrading in various industries and more accurate and efficient screening, convergence, analysis, and prediction of massive situational perception information. The intelligent digital twin connects the data of cloud, edge and terminal layers of the platform, and processes and analyzes the data. It has the capabilities of holographic mapping, simulation deduction, analysis & prediction and real-time interaction. The full-stack technology system can quickly adapt to various industries, meet the cost reduction and efficiency increase requirements of business digital transformation, and provide strong support for establishing industrial upgrade with data as the core driving factor.

BSS can adopt a zero trust security model to ensure security. In essence, zero trust is to authenticate all assets such as personnel, devices and servers and build an identity centric trust evaluation and dynamic access control system to ensure security protection for business data access. That is, through the zero trust business access model, it ensures that the correct people, use the correct terminals, use the correct authority to access the correct business at any network location and obtain the correct data. A decentralized trusted identity authentication solution based on blockchain can be applied to identity authentication and manage keys in decentralized related business scenarios.

BSS can also introduce low code platforms, which can enhance communication efficiency between business teams and IT departments by reducing losses caused by human communication. For developers, low code platforms eliminate the tedious and repetitive coding work in the development process, and can effectively reduce labor costs and improve the development efficiency. In terms of low code, the super APP solution adopted provides a very convenient service. With strong third-party integration capabilities, it offers multiple services by a single application and can save resource and realize data sharing. On this basis, a super automated business technology platform is constructed, which is composed of a business automation orchestration module, a data governance module, a data flow orchestration modeling module, an AI algorithm module, a visualization module and a SmartAgent module. For scenarios where process standardization rates can be applied, the platform can comprehensively solve manual errors in business processes and improves enterprise process efficiency.

## 5.2 6G BSS Key Technologies

### 5.2.1 Total Experience Upgrade Based on Intelligent Customer Avatars

Undoubtedly, total experience [35] will become one of the main options for communication operators to move towards Web3.0 metaverse operation based on 6G capabilities. It is a business strategy implemented for all enterprise related personnel (including customers, employees, users, partners, etc.), with the goal of creating a more comprehensive and more pertinent brand perception and experience for everyone. Total experience includes four categories: customer experience, i.e. to measure customers' perception and experience of the brand through their interaction and behavior with the enterprise; employee experience, i.e. to measure employees' perception and experience of the brand through their interaction and behavior with customers; user experience, i.e. to measure the users' experience of this application through their behavior and feelings during the use of digital products; multiple experiences, i.e. to measure the experience of different interaction modes brought to users by the same application across devices and contacts.

With the upgrading of computing and network infrastructure brought by the 6G network, the customer avatar [36] based on Web3.0 is expected to significantly enhance the total experience of the metaverse as a killer application: customer avatar, as customer's digital avatar, uses the key of an Internet token to smoothly navigate different applications, while enterprise employees can also verify the identity of the same user, both parties have established a mutual trust relationship based on trusted technology; customer avatar provides customers with personal data management throughout their full life cycle, including authorizing third parties to use data and personal data transaction. Enterprises obtain legal and compliant data utilization rights through smart contracts, and employees do not need to worry about compromising their personal privacy; customer avatar provides an open interface that allows customers to purchase and access intelligent services provided by enterprises, enabling them to act as intelligent assistants to help customers effectively manage the interaction behavior and time of the metaverse. In terms of user experience and multiple experiences, regardless of the device or terminal, customers can interact with their main or even unique contact customer avatar in a natural way. For example, a virtual reality helmet can be operated through voice intonation, eye movement, facial expressions and even body language, while wearable watches, allows user to interact through voice, or text, in an environment that requires silence. In short, all parties in the metaverse communicate effectively by natural language interaction at different contacts through customer avatar. Customer avatar can always be online, react in real time, and continuously optimize itself.

In the metaverse era supported by 6G network, customer journey will become the most fundamental and core element, and effective management and sharing of customer journey will become the cornerstone of continuous optimization and total experience [37]. The characteristics of current customer journey data aggregation are that data needs to be injected in a real-time, agile and continuous manner, and data from different systems and sources need to be matched based on the unified identity of the customer. In the customer journey analysis phase, the focus is on connecting customer behavior trajectory data from different channels, contacts and interfaces based on time series to reveal the customer's true history trajectory, replacing the original scattered analysis of contacts with linear trajectory path analysis. In terms of the application of customer journey strategy, it can simultaneously bring benefits in the three dimensions of enterprise value beliefs (product leadership, customer intimacy and operational excellence), focusing on four aspects: operation efficiency enhancement, experience improvement, revenue growth and customer loyalty. With customer avatar as an innovative interface for user experience, customer-centric, immersive and wrap-around integrated contact services will become the mainstream in the future. The granularity of contact, communication and interaction between enterprises and customers will become increasingly detailed, covering a wider customer journey and a more extensive multi-dimensional service space.

## 5.2.2 Upgrade of Interaction Methods Based on Intention Perception

The technological development in the 6G era will be inclined to interact with systems with natural ways in order to improve efficiency and lower the threshold for system use, and enable the intelligent agents to obtain more training data for faster iterations. Therefore, the interaction method based on natural input/output modes such as voice, coupled with intention perception[38], will definitely promote the upgrade of interaction method.

Intention perception-based interaction is an intelligent interaction method, by which the user's language, semantics, background and other information can be analyzed to understand the user's intentions and provide corresponding help or answers. This interaction method can make the user experience more natural, and help users obtain the required information more quickly and accurately. Compared to traditional interaction methods, intention perception-based interaction can provide more personalized and flexible services and more efficient solution to problems. Implementing this interaction method requires strong natural language processing capabilities and a rich knowledge base; meanwhile, issues such as uncertainty, ambiguity and vagueness, as well as how to improve the accuracy and credibility of the system also need to be addressed. Intention perception-based interaction can be combined with voice as an input/output method. This allows users to interact with the system by speaking without using text or other input methods.

Using BSS based on intention perception, voice input and other technologies can achieve rapid upgrades in interaction methods, greatly reduce the threshold for system use, and enable end customers, temporary employees and others to use the system efficiently and accurately. In addition, it can also improve production efficiency, enable users to obtain more accurate business operation data faster, and make timely and accurate responses to the market.

### 5.2.3 Visual Operation Based on Intelligent Digital Twins

Due to the fully open architecture of BSS in the 6G era and the unprecedented enhancement of communication network capabilities, a large number of advanced technologies will be integrated into the system, and BSS will penetrate all aspects of the cloud, network, edge and terminal, all of which lead to the unprecedented complexity of BSS. At the same time, 6G has carved out many new frontiers for the development of telecommunication business, and has also made business development and market competition more complex and difficult to control. Visual operations based on intelligent digital twin can reflect the architecture and operational status of the system, as well as changes in business in an intuitive and real-time manner, and enable a certain degree of prediction and simulation for business.

The intelligent digital twin [39] technology allows entities or processes in the physical world to be digitally mirrored and replicated in the digital world. Intelligent interactions between people and people, people and objects, objects and objects may be achieved through mappings in the digital world. By mining rich historical and real-time data in the digital world and generating perceptual and cognitive intelligence through advanced algorithmic models, the digital world can simulate, verify, predict and control physical entities or processes, thereby obtaining the optimal state of the physical world. Intelligent digital twin technology can provide many possibilities in various fields: in the medical field, medical systems can duplicate the patient information via digital twin to make disease diagnosis and determine the best treatment plan; in the industrial field, optimizing product design through the digital domain can reduce costs and improve efficiency; in the agricultural field, using digital twin to simulate and deduce agricultural production processes can predict adverse factors and improve agricultural production and land use efficiency; in the field of network operation & maintenance, through closed-loop interaction between digital and physical domains, cognitive intelligence, automated operation & maintenance and other operations, the network can quickly adapt to complex and dynamic environments, thus achieving "autonomous" throughout the full life cycle of operation & maintenance, such as planning, construction, monitoring, optimization and self-healing.

The application of intelligent digital twin technology in BSS can include two aspects: system twin and business twin. In terms of system twin, BSS can use twin technology to construct their own operation & maintenance system: for example, twin technology can be used to reflect the system architecture and system operation status in real time, thereby detecting system operation problems easily; the twin can also be used to simulate some extreme scenarios (such as large-scale sports events held in cities), so as to predict potential problems for the system operation under such scenarios and explore better system resource configuration for higher operation performance or reduced system costs. In terms of business twin, digital twin technology may be used to enhance the observability of applications and provide real-time analysis and intuitive presentation capacities for actual data generated during enterprise operations. Digital twin technology may also be used to simulate business decision results, predict the effects of marketing plans, etc., facilitating proactive and informed decision-making of operators in a relatively short period of time. As a result, operators can further meet the customer-centric digital transformation needs and enhance the competitiveness of enterprises.

### 5.2.4 Automated Operations Based on Self-adaptive AI

Due to the unprecedented business and technical diversity in the 6G era, BSS has become more complex, and the requirements for availability and reliability are more stringent. In this context, maintenance of BSS only with manpower would be unfeasible, thus making AI technology indispensable for BSS operations. However, the current AI technology still requires a large amount of human intervention to achieve feature engineering, parameter adjustment and even algorithm adjustment based on changes in business scenarios. This will also restrict the adaptive speed and ability of the system in the 6G era of rapid change. Therefore, the introduction of self-adaptive AI technology will enable BSS in the 6G era to have better business adaptability, thereby helping operators gain advantages in competition.

Self-adaptive AI [40] refers to the ability of an AI system to automatically adjust its parameters and behavior based on the current environment and tasks to better complete tasks. This type of AI system can remain effective in ever-changing environments and continuously improve its performance through learning. In this case, the AI system does not need human intervention, and can learn and adjust independently to adapt to the new situation.

The self-adaptive AI technology can improve the operation efficiency and performance of BSS. This technology can automatically monitor the operating status of BSS and automatically adjust parameters based on the system needs and performance. For example, when a system requires more computing power, the self-adaptive AI technology can automatically expand the computing resources of the system; when the system is operating normally, the operating parameters of the system can be automatically adjusted to save energy and improve performance. In addition, the self-adaptive AI technology can also continuously improve the system performance through learning, enabling the system to respond to ever-changing environments and needs.

### 5.2.5 Distributed Application Based on Cloud-edge-terminal Integration

In the 6G era, the support for business will show decentralized characteristics, and the information collection, computing, decision-making, marketing, and service activity operations of business operations will not only be executed in the center system. Therefore, BSS will not only be deployed and executed in the cloud, but also on the edge and terminal.

Therefore, it will be necessary to utilize cloud-edge-terminal integration technology to support the operation & maintenance of BSS as a distributed application.

The cloud-edge-terminal integration technology [41] achieves unified perspective management and use of cloud-edge-terminal distributed resources by integrating and abstracting edge nodes and terminal devices with wide distribution, heterogeneous resources, diverse forms and different protocols. On the basis of resource integration, it is possible to uniformly monitor, operate & maintain cloud-edge-terminal distributed resources, perform unified perspective operation & maintenance, and minimize user operations, thereby building a full range of capabilities and three-dimensional security capabilities for distributed data collection, processing, aggregation, analysis, storage and management.

BSS based on cloud-edge-terminal integration technology can provide unified application and operation management for business scenarios by integrating capabilities such as ubiquitous access, network management, cloud-edge-terminal collaboration, unified scheduling, AI, data platform, component development and ecological openness, shielding differences in underlying heterogeneous resources. With the support of cloud-edge-terminal integration for BSS, operators can screen, converge, analyze, predict and respond to massive information more accurately and efficiently, thereby effectively improving business agility and responsiveness.

## 5.2.6 Native Network Security Based on Blockchain Trusted Identity Authentication

After entering the 6G era, the related BSS will be provided and operated by operators, while partners and customers may become system providers and operators. The network environment is not limited to the operator's intranet, and the operating device is not limited to the operator's device, which will become inevitable. Therefore, it will be inevitable to ensure the high security of BSS based on Native Network Security concepts and technologies, and the blockchain-based trusted identity authentication technology [42] is the key to achieving Native Network Security.

In the traditional identity authentication system, the identity information is stored and managed by the central server. It is common for authentication servers to experience hardware and software faults and network attacks. In order to thoroughly address the security threats of identity authentication systems, it is urgent to find a decentralized technical solution that has high data security, effective resistance to network attacks, and can meet the operation needs of authentication systems. A decentralized identity is a new form of identity and access management (IAM), which is no longer the centralized storage of user information. Decentralized identity identification stores identity information in distributed computer systems, such as distributed ledgers or blockchains, and uses distributed ledgers to preserve identity elements to protect them from tampering and theft. Therefore, even if your identity information is recorded in electronic format, it will not be changed, stolen or deleted.

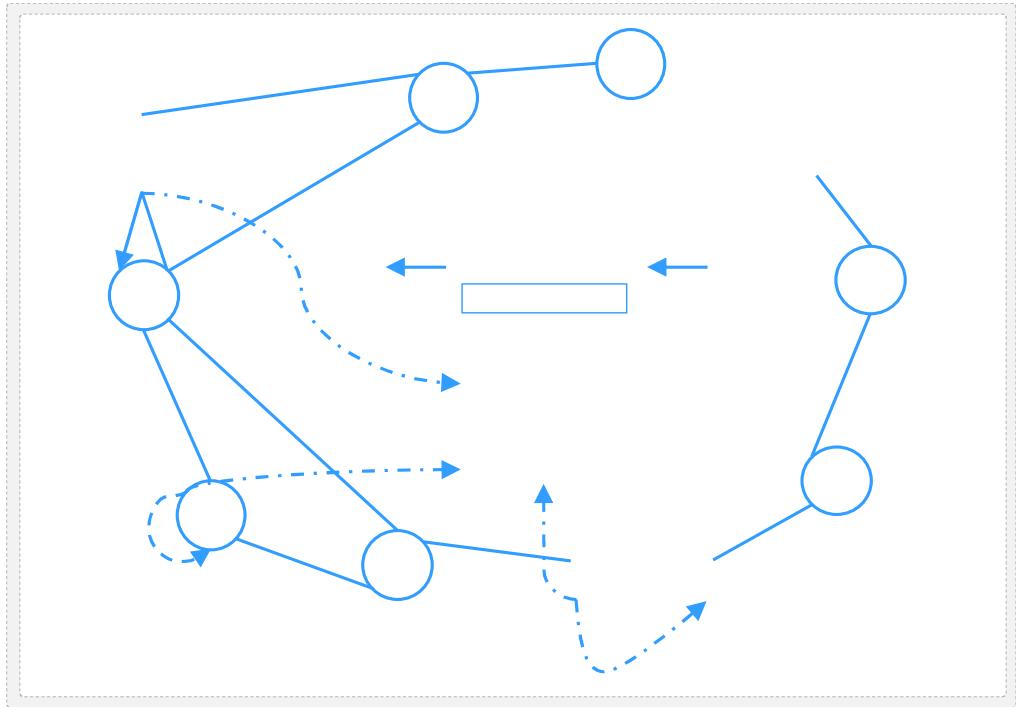

**Figure 5-3 Decentralized Identity Solution Based on Blockchain**

After BSS adopts a blockchain-based identity authentication solution (as shown in *Figure 5-3*), user nodes only need the IGC (Identity Generator Center) to issue identity information and generate key information when registering for the first time. Subsequently, user nodes can automatically update their keys without relying on the IGC, and the key update process is recorded on the blockchain in the form of transactions. The immutability of blockchain data ensures the security and credibility of the key update process, thereby achieving high security of BSS in identity authentication and key management.

## 5.2.7 Minimalist Development Based on Super Automation and Platform Engineering

6G BSS faces unprecedented pressure of business operation and upgrade maintenance. How to cope with these pressures with lower costs and faster response is an important test for BSS, while super automation and platform engineering are the key technologies to address this pressure.

Super automation [43] is a business-driven method, which is used by organizations to quickly identify, review and automate business and IT processes as many as possible. It is also an integration combining multiple technical capabilities and software tools such as robotic process automation, process mining and intelligent business process management, a further extension of concepts such as intelligent process automation and integrated automation, and an important part of enterprise digital transformation, remodeling and revolution. Business-driven super automation, as a new methodology, achieves collaboration, reshaping and optimization of business processes across departments and systems in enterprises through a lightweight, non-invasive method. The super automation solution makes it possible to quickly identify, review, automate business and IT processes as much as possible, and enables users to expand and accelerate the digital process. Super automation has realized the automatic processing of massive complex businesses and has been widely used in industries such as finance, manufacturing and telecommunication, effectively promoting the digital transformation of organizations. With this concept, BSS can be constructed as a super automated business technology platform, which can be applied in scenarios with process standardization rates to comprehensively solve human errors in business processes and improve the efficiency of enterprise processes.

Platform engineering technology is a set of mechanisms and architectures for building and operating a self-service internal developer platform that supports software delivery and life cycle management (as shown in *Figure 5-4*). The goal of platform engineering is to optimize the developer experience and accelerate the product team's speed for creating value for customers. The integrated products provided by the platform engineering are commonly referred to as "internal developer platforms", covering the operation & maintenance needs throughout the full life cycle of an application. According to Gartner, by 2026, 80% of software engineering organizations will establish platform teams, and 75% of those will provide self-service developer portals.

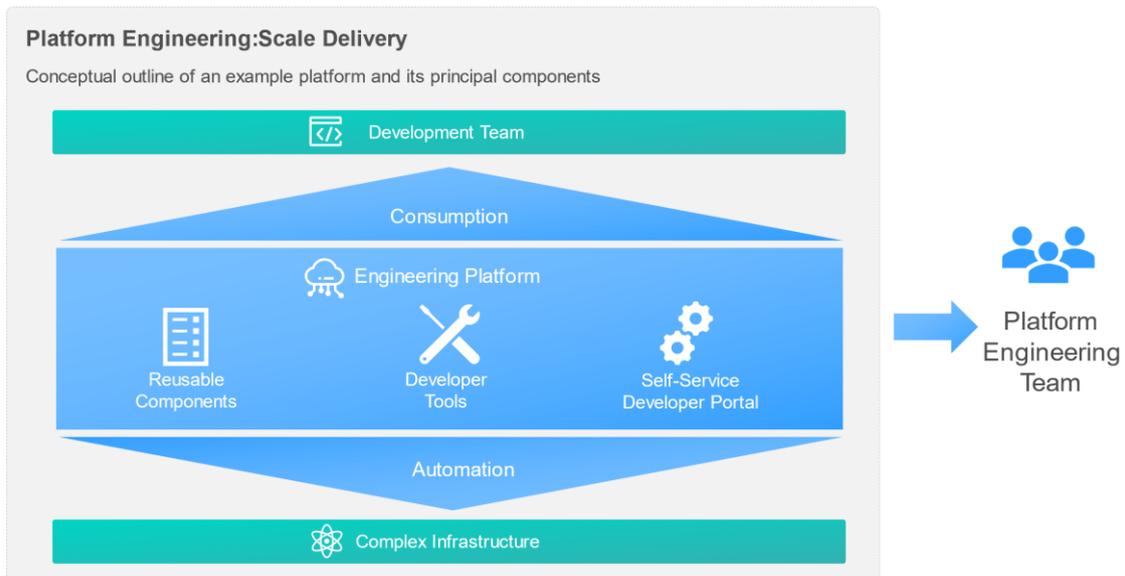

**Figure 5-4 Platform Engineering Technology**

Platform engineering achieves a certain level of developer self-service through product methods, and finds the appropriate abstraction level for various organizations and teams. Platform teams can combine user research, regular feedback, and best marketing practices to get to know their developers, create a platform to solve common problems, and gain internal support from key stakeholders.

BSS of the application platform engineering can minimize the resistance encountered by developers when completing daily tasks. Recommended tools and best security practices are also provided to reduce the cognitive burden of developers, while retaining a certain degree of freedom. All of these efforts have ensured that the platform can reduce the cognitive burden and achieve an appropriate balance between developers' needs for self-service and support.

## 5.2.8 Automatic Operation & Maintenance Based on Digital Immune System

Due to the diversity in system composition, business and technology of BSS in the 6G era, the complexity has increased, and the difficulty of preventing and repairing faults will increase exponentially compared to the 5G era. Therefore, it is very important to introduce digital immune system technology to achieve automatic operation & maintenance.

Digital immune system (DIS)[44] combines practices and technologies such as observability, AI enhancement testing, chaos engineering, self-healing, site reliability engineering and software supply chain security to improve the flexibility of products, services, and systems while reducing business risks (as shown in *Figure 5-5*). A powerful digital immune system can make applications more resilient and quickly recover from faults, thereby protecting applications and services from abnormal impacts, such as software faults or security issues. The digital immune system can reduce business continuity risks when critical applications and services are severely damaged or completely stopped. Enterprises face unprecedented challenges in ensuring a resilient operating environment and accelerating digital delivery and reliable user experience, and they want to respond to market changes and rapid innovation quickly. Users expect not only perfect functions, but also high performance, transaction and data security, as well as satisfactory interaction.

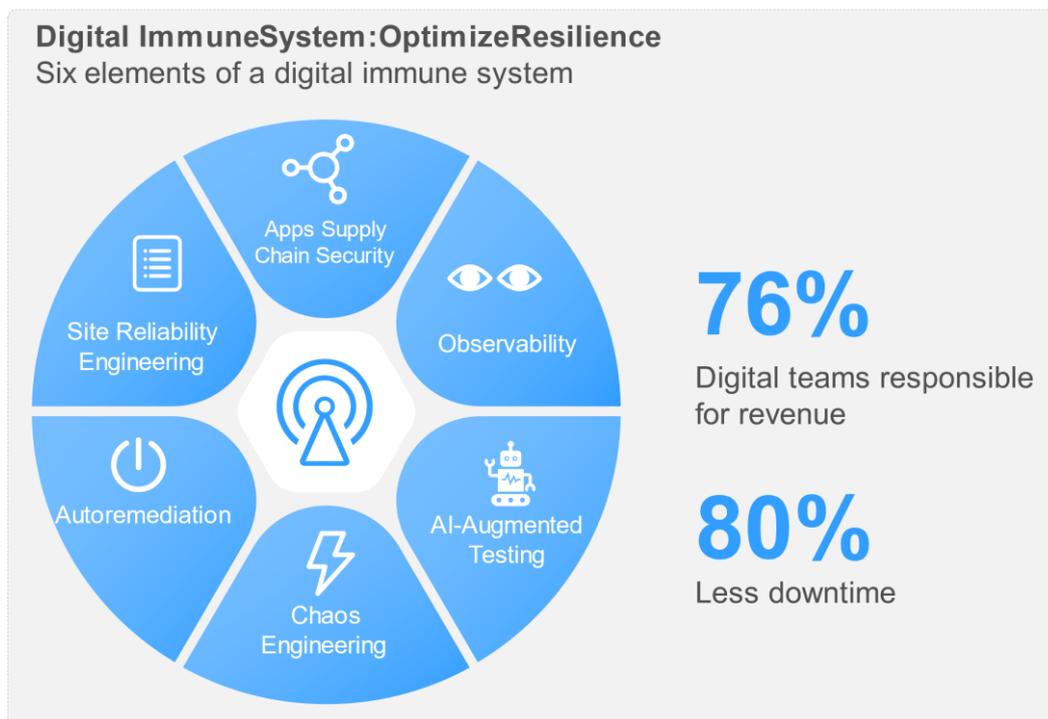

**Figure 5-5 Digital Immune System Technology**

BSS using a digital immune system will create an excellent maintenance experience while reducing system failures (research has shown that it can reduce system downtime by up to 80%), enable the system to quickly recover from faults, and directly convert the reduced losses into higher revenue.

## 5.2.9 Data Capitalization Based on Data Fabric

As comprehensive information service providers in the information society, communication operators have the natural advantage of data pipelines. The information fingerprints of people in the modern society are recorded in detail in their network systems and business platforms. Communication operators have accumulated data for many years, and have five characteristics (namely "true, large, fast, flexible and complete") in data resources. Among them, "true" refers to the authenticity and high accuracy of operator data; "large" refers to massive operator data covering core production operations, customer contact, customer behavior perception, network elements, etc.; "fast" refers to high timeliness of operator data; "flexible" refers to scalability, loose coupling and flexible opening of operator data; and "complete" refers to the coverage of structured, semi-structured, unstructured multiple data types and multiple data sources. The data of telecom operators involves all the businesses including mobile voice, fixed phone, fixed network access and wireless Internet access, etc. as well as public customers, government & enterprise customers, and household customers. Moreover, contact information from all types of channels such as physical channels, electronic channels and direct sales channels will also be collected. The development of 6G will further enhance the above features, such as its joint communications and sensing ability with native intelligence, which will greatly improve the speed, accuracy and information value of data acquisition. However, in order to meet the large-scale connection, ultra-low latency communication and diversified customization requirements in the 6G era, distributed deployment of 6G networks will become inevitable. Meanwhile, due to security and privacy considerations, enterprise users have high requirements for the privacy and confidentiality of their production and business data, and there is a strong demand for data not leaving the plant. This requires that the network not only achieves local transmission and analysis of data, but also achieves local transmission processing of control signaling. Therefore, how to effectively manage distributed data assets and assist operators' lean operation will be an important issue in 6G BSS.

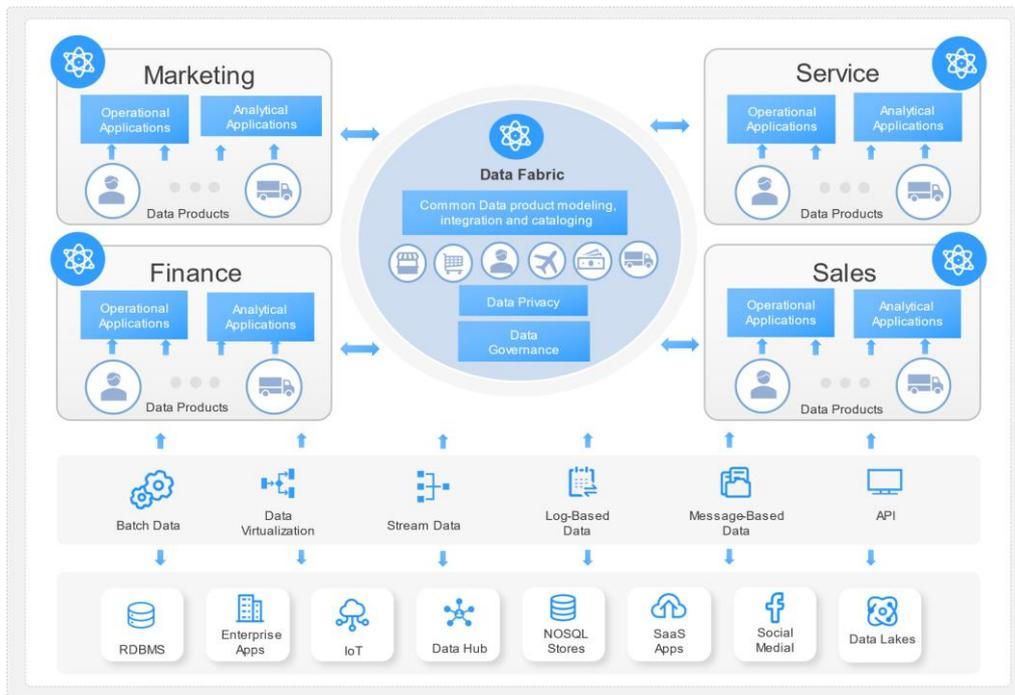

**Figure 5-6 Data Fabric Technology**

Gartner defines Data Fabric as an integration layer that includes data and connection. Through continuous analysis of existing, discoverable and inferable metadata assets, Gartner supports the design, deployment, and use of data systems across platforms, thus enabling flexible data delivery. Data virtualization is one of the key technologies in implementing a data fabric architecture. It can access data from the source without moving it and can shorten the time to achieve business value through faster and more accurate queries. Specifically, it includes functions such as cross-platform agile integration, unified semantics, creation of data APIs (supporting technologies such as SQL, REST, OData, and GraphQL) with low code and intelligent cache acceleration. Data fabric can be taken as a virtual network [45], where each node is a data source. This network represents a virtual connection that allows data to flow quickly online and provide unified external services (as shown in *Figure 5-6*).

The distributed deployment of 6G networks results in data assets being hidden in a mixed combination of infrastructure environments. Due to the long data preparation cycle, users need a wide range of data management functions to overcome complex constraints such as multi suppliers, multi-cloud and evolving data environments. Data fabric technology is designed to address the challenges of complex mixed data environments. Essentially, the data fabric supports various data management requirements to provide the correct IT ser-vice levels across all different data sources and infrastructure types. It operates as an integrated framework for deploying, managing, moving, and protecting data across multiple isolated and incompatible data centers. As a result, organizations can invest in infrastructure solutions that meet their business needs without worrying about data service levels, access and security.

## 5.2.10 Maximizing Data Value Based on Privacy Computing

Operator data is valuable not only for operators' business operations but customers in other industries. Operator data has many valuable application scenarios in industries such as finance, tourism, education and automobile, and can output value to customers in these industries. Taking the financial industry as an example, the payment information, location information and interest preferences of operators' customers can bring greater value to the business operations of customers in the financial industry. According to the true, accurate, and sustainable data resource supply of operators, financial institutions can further enrich and supplement the labels and portraits of financial customers, assist financial customers to gain in-depth insight into user's behavior and interest preferences, and can play a role in scenarios such as identity verification, credit reporting, fraud prevention, precision marketing and site selection. The application value of operator data of the 5G era in other industries has been widely and fully verified in practice.

How to ensure that operators can achieve deep data convergence and data value release across organizations while ensuring data security and compliance? Privacy computing[46] provides a perfect solution to solve this problem. Privacy computing is a computing theory and methodology oriented to the full life cycle protection of private information as well as a computable model and axiomatic system for privacy metric, privacy leakage cost, privacy preservation and privacy analysis complexity in case the ownership, management and utilization rights of private information are separated (as shown in *Figure 5-7*). Via privacy computing, the balance and common development among "data silo interconnection", "data privacy preservation" and "business development" can be achieved. Privacy computing can realize the flow and sharing of data "value" and "knowledge", and can significantly improve the efficiency of data cross-domain convergence. At present, it is helping operators to achieve faster data cross-domain monetization.

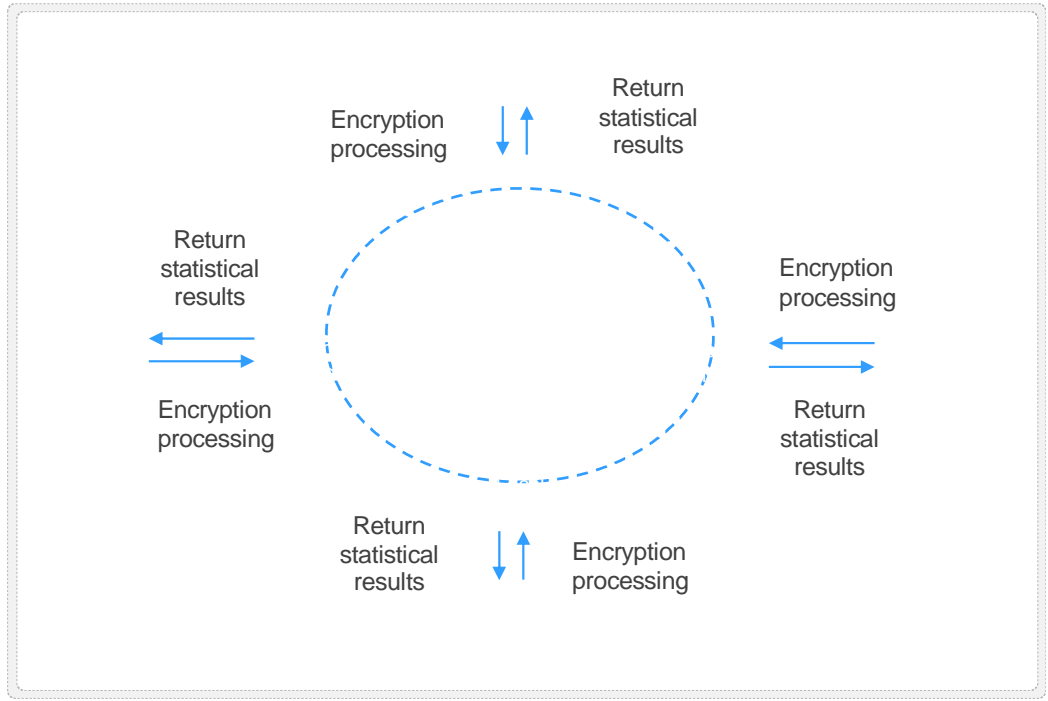

**Figure 5-7 Schematic Diagram of Secure Multi-party Computing**

Privacy computing is a general term for the technical routes to solve the problem of information confidentiality in computing links during data circulation. It may be defined as a kind of technical set or system that realizes data analysis, computing and application without raw data transmitted or with raw data preserved. From the perspective of technical mechanism, there are mainly three technical routes for privacy computing, namely, secure multi-party computing based on cryptography, federated learning based on distributed training and computing environment based on trusted hardware. With the development of Native Network Security and Native Intelligence under the further convergence of 6G computing and networks, operators can provide efficient and reliable privacy computing capabilities for releasing their data value by establishing trusted computing nodes with secure multi-party computing or federated learning (as shown in *Figure 5-8*).

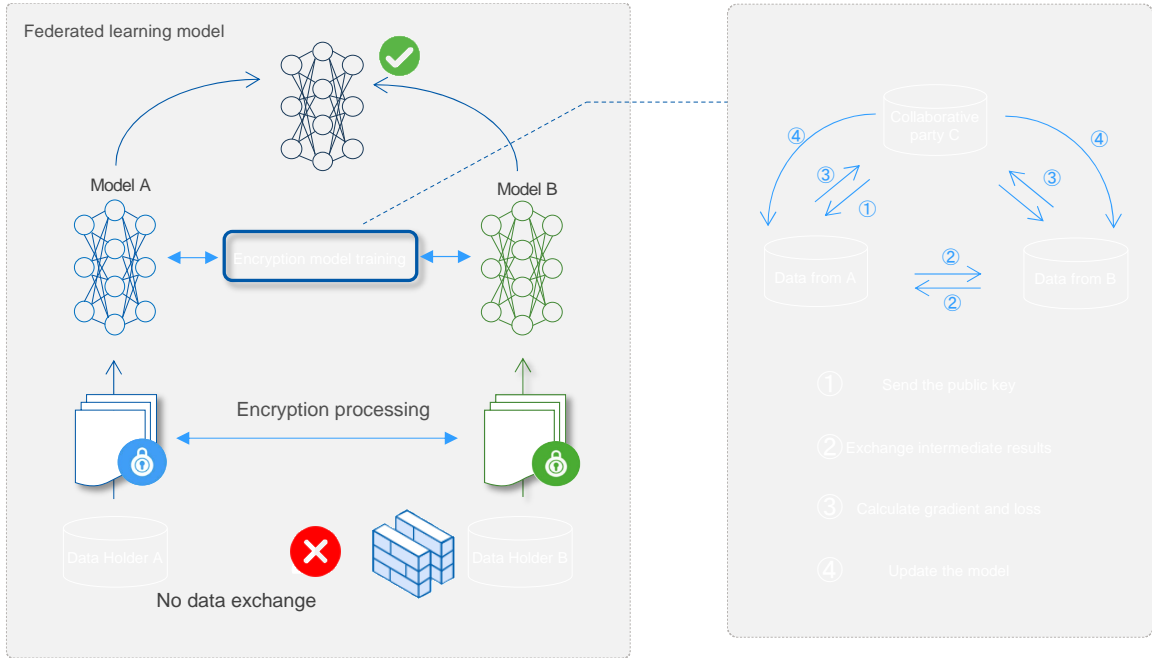

**Figure 5-8 Schematic Diagram of Federated Learning**

## 5.2.11 PaaS Capability Evolution Based on Sustainable Cloud Computing Model

PaaS is an intermediate layer in cloud computing. In many people's eyes, PaaS layer is less closely related to green & low carbon than the IaaS layer. However, PaaS service can better and more efficiently schedule and manage IaaS resources; establish common services to provide service reusable for SaaS-based applications, and help applications achieve rapid innovation and reduce development volume, reduce operating expenses and save energy costs, which are contributions made in term of the sustainability of PaaS service in IT cloud service[47].

The ecological sustainability of PaaS service[48] is currently in the early stage of exploration. Based on the own practice of BSS, the following key technologies can be on trial:

- **Virtualization technology**

    Compared with the traditional virtualization technology at the IaaS layer, the container technology widely used at the PaaS layer can significantly improve the utilization rate of resources. Containerization technology reduces the use granularity of resources from level GB to level MB, which makes the application and release of resources more accurate, thus reducing the invalid input of resources. At the same time, virtualization technology at the PaaS layer can directly run assembled applications on physical servers, which further reduces

the performance overhead incurred by IaaS layer virtualization by about 30%. By reducing the number of servers with virtualization technology, IT teams will reduce the maintenance time of physical hardware and IT infrastructure, thus improving team efficiency and productivity. At the same time, fewer physical servers adopted will directly reduce energy consumption[49].

- **Multi-level resource pool monitoring, scheduling and recycling**

  After the introduction of container technology, multi-level resource supply and scheduling will be formed at the PaaS layer. The IaaS layer will provide basic computing and storage resources to the PaaS layer, and form a first-level resource pool at the PaaS layer, some of which will be used to provide basic resources for containerization technology. The container resource pool will provide secondary resources to different tenants. When tenants apply for resources, in order to ensure the stability of the business, the resources will be generally allocated according to the peak usage of the business. According to the actual situation during the operation, the peak usage of business only accounts for a small part of the total operation time. This means that the resources allocated for the peak are wasted most of the time. Through monitoring, scheduling and recycling with multi-level resource pools, the idle resources of tenants can be recovered to the secondary resource pool for tenants, and through real-time monitoring, the resources can be supplied to tenants in time when their business volume increases. In this way, the overall resource utilization rate at the PaaS layer will be improved, and the resource demand will be reduced accordingly, thus reducing the resources consumption at the IaaS layer, the server investment and the corresponding cooling energy consumption investment, and improving the sustainability.

- **Common capability reuse**

  In recent years, the construction of the Middle Platform in China has been in full swing in industries with successful digital transformation. The core of the Middle Platform strategy is to capitalize IT capabilities and enhance the reuse capability of IT assets. Relying on PaaS platform, construct atomized microservices and provide life cycle management for microservices. By capability exposure, building a large PaaS ecosystem, empowering SaaS business development, and gradually transforming business coding into capability arrangement, the business development efficiency is greatly improved, and various manpower and material resources incurred by repeated development will be reduced, thus lowering carbon emissions and improving ecological sustainability.

● **Cross-cloud dynamic scheduling technology**

Because the underlying layer supports and manages the resource pools of different cloud data centers, it is necessary to support the scheduling of different resource pools at the underlying layer when the PaaS layer is deployed or flexibly expanded, which involves the opening of possible heterogeneous resources and networks. On the scheduling algorithm, the perception capability of renewable resources shall be supported, for example, sensing that the data resource pool of a cloud service provider is green and renewable, while the resource pool of another service provider uses traditional fossil fuels. In order to improve ecological sustainability, it is necessary to identify the resource types in the scheduling algorithm, so as to achieve the best scheduling scheme in line with sustainability.

## 5.2.12 Cloudified Architecture Based on High Performance Computing

High performance cloudified architecture is a high performance computing service model combined with cloud computing technology, with high performance computing as the core of service, cloud computing as the technical means of service model innovation, and multi-cloud interconnection as the scalable support of service capability. On this basis, the HPC Cloud will be deeply integrated with technologies such as big data and AI to provide integrated intelligent computing service capabilities and realize the scalability of HPC Cloud capabilities oriented to the application needs of the industry.

To realize BSS high performance cloudification[50], the followings shall be included in operation management, architecture, technology and operations management:

● **Operation management**: Integrated operation

The cloudification of BSS is to improve operational efficiency and reduce operating costs, especially IT costs. At present, telecom operators adopt partition-based and region-based operation management modes. Different services fixed network, (mobile network and broadband) are operated by different business units, and different regions are operated by different subsidiaries or subnets. However, in this way, it is difficult to exploit the collaboration advantages, resulting in low overall operational efficiency and poor service experience for the whole network. So it is necessary to integrate cross-department business processes and realize the integrated operation of different subnets.

The changes in operation mode and business process involve management reform, which is a long-term process and needs the firm determination of operator decision-makers to realize. Integrated operation not only improves operational competitive strength and efficiency, but also lays a foundation for the maximum sharing of IT resources from hardware to application software. From the aspect of business, the differentiated support by cloudification of BSS for operation is a key issue, which must be supported by multi-tenant mode via full servitization with business functions, flexible business rules and process management.

- **Application architecture:** Separation of functional and non-functional characteristics

In terms of application architecture, it must completely isolate the functional characteristics (business logic) and non-functional characteristics (deployment, performance, reliability, scalability and maintainability) of software through the business platform layer during the cloudification of BSS, to realize the dynamic deployment mechanism, on-demand dynamic scalable mechanism and self-healing mechanism based on SLA without increasing the complexity of application logic.

That is to say, SLA may not be considered when realizing presentation logic, function logic, data storage and access logic of business function, and SLA protection is completely handled by the business platform layer. According to the requirements of SLA, including response time, processing time limit, allowable interruption time, the business platform layer deploys the application in a suitable hardware environment (for example, cluster deployment shall be used for high availability) and realizes dynamic scalability via the real-time monitoring of SLA. For example, if the business platform layer finds that there is a lagging trend for the response time of the application with the increase of load, it will allocate a new host from the resource pool, conduct on-demand deployment for the application, and then modify the route to allocate the extra load to the new host, maintaining the SLA unchanged. When the service platform layer finds that the load processed by the host drops to a certain extent, it will modify the route, accumulate the load on a small number of hosts, release the idle hosts, unload the application deployment, and put it back into the resource pool.

For another example, if the business platform layer found that with the increase of data volume of the data sheet, the data access speed decreases, resulting in an overall SLA decrease, at this time, the system will dynamically split the data in the data sheet, distribute them to multiple databases, modify the data route, and allocate the access request to multiple databases, thus maintaining SLA. By the same mechanism, when faults occur at different levels such as database, host and storage, the application can automatically recover from the faults and maintain the stability of SLA.

In order to meet the requirements of quality of service, the hardware configuration of traditional BSS usually meets the peak processing load, which is 2 ~ 3 times the average processing load. At the same time, traditional BSS adopts static application deployment mode, so even if the peak time of different applications is different, it is impossible to balance the use of hardware resources by making use of the staggered peak time. The above factors lead to the low overall utilization rate of hardware, with an average utilization rate of about 30%~40%. Technically speaking, after realizing the dynamic deployment of applications, the utilization rate of hardware resources can be improved and the cost of IT construction can be reduced. For example, the idle resources of the settlement system during the day can be released and used by CRM and billing systems when they are in high traffic demand.

- **Technical architecture:** Separation of business and technology

In terms of technical architecture, the business platform layer will completely isolate business functions from infrastructure (hardware, middleware, database, etc.). The engineering of service functions does not depend on specific technologies, and the service platform layer adopts fully distributed application and data architecture. So operators can choose appropriate technologies to meet service requirements from the comprehensive consideration of cost and efficiency. Traditional BSS is often bound with specific technologies during design, for example, adopting centralized architecture suitable for minicomputers instead of distributed architecture, so it is impossible to adopt blade servers with lower cost conveniently.

For another example, traditional BSS generally adopts centralized database architecture. With the increase in users and transactions, centralized databases will gradually become the bottleneck of system reliability and performance. Moreover, a small problem with the centralized database will affect the service experience of all customers. The architecture of a traditional BSS system strongly depends on a centralized database. If being changed to distributed database architecture, the application transformation workload is huge and the feasibility is little. Therefore, the prime balance between cost and quality of service may be achieved only by completely separating business functions from the infrastructure layer in the architecture.

- **Operations management:** Automation and intelligence

  Cloudified BSS needs to realize intelligent automated management of operation & maintenance. Traditional manual maintenance process leads to frequent business interruptions and human errors, for example, simple system expansion requires long-term business interruptions, which affects the quality of service for customers. Only by setting rules in advance to realize automated upgrades and fault correcting, can the quality of service for customers be improved.

## 5.2.13 Intelligent Ubiquitous Scheduling of Computing Power Based on Different Hardware Platforms

In the 5G era, due to the gradual migration of communication network infrastructure from a dedicated hardware platform to a general hardware platform, basic hardware based on architectures such as x86 and ARM has been widely adopted, and computing power constructed mainly based on heterogeneous chips such as GPU, NPU and FPGA will become the global mainstream in the next 5~10 years. This trend puts forward higher requirements for intelligent  ubiquitous scheduling of computing power based on different hardware platforms.

To realize the ultimate performance release of intelligent computing power, in the architecture of 6G BSS, it needs to consider the horizontal collaboration of computing, network and storage, and the vertical collaboration of software platform and hardware resources, so as to support the continuous enrichment of intelligent business application scenarios of 6G. Such demand promotes the development and innovation of intelligent computing power ubiquitous scheduling technology to meet 6G and future business requirements. For example, the XPU chip platform and its software tools (such as OpenVINO, BigDL and Analytics Zoo) based on Intel Xeon scalable processor accelerated by built-in AI can help realize computing-aware scheduling and optimization across different hardware platforms, and also meet the computing and intelligent processing requirements of 6G and future business requirements. On such a platform, intelligent computing power can be fully utilized, and can also help promote the continuous development and innovation of intelligent application scenarios. Therefore, intelligent ubiquitous scheduling of computing power based on different hardware platforms has become one of the key technologies to realize 6G and future business applications. This technology will help improve the stability, security and flexibility of the network, so as to meet the changing business requirements and consumer requirements.

# VI. 6G BSS Potential Architecture Evolution

## 6.1 6G BSS Design Methodology

Operation management capability, including a series of activities such as planning, organizing, commanding, coordinating and controlling the production and operation activities of enterprises, is the core competitiveness of enterprises and plays a critical role for all enterprises. Enterprise digital intelligence is to digitize business activities by scientific and technological means and provide managers with all information for observation and insight into various matters, based on which the managers will be able to make rational management decisions, optimize the allocation of resources and enhance economic benefits.

6G BSS system is an effective tool to improve operators' digital intelligent operation and management capability. It is of critical importance to know how to plan its core functional architecture model and realize the closed-loop management of the enterprise. PDCA, as a kind of management thought, was first put forward by American quality management expert Walter A. Shewhart, and then adopted, publicized and popularized by W. Edwards Deming, so it is also called Deming Cycle (as shown in ). From the planning and continuous improvement of quality management activities to the operation management field in all walks of life, Deming Cycle becomes the thinking foundation and method for enterprise comprehensive operation management. The core is the PDCA cycle, which divides the quality management of enterprise operation into four phases, namely Plan, Do, Check and Act, which requires all the work to be planned, the plan to be implemented and the implementation effect to be checked, with the successful ones being included in the standard, and the unsuccessful ones being left to be solved in the next cycle[51, 52, 53, 54, 55].

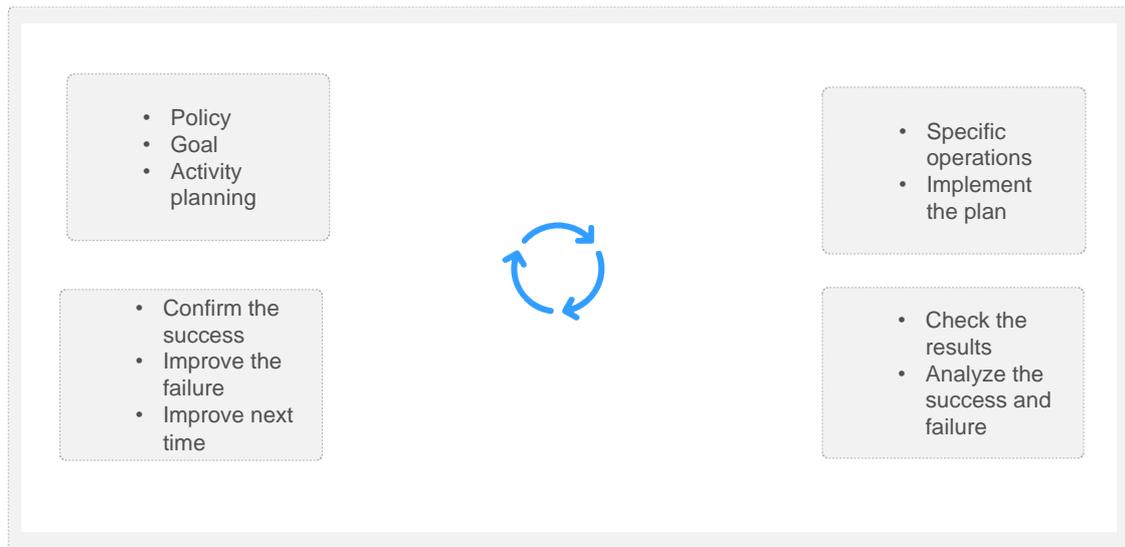

**Figure 6-1 PDCA Operation Management Model**

As a new generation of business innovation platform, 6G BSS can take PDCA as the core operation management model to realize PDCA architecture via digital intelligence technology, comprehensively covering operators' staff, property and belongings as well as production, supplying and marketing, and building a complete management cycle from strategy to implementation; in addition to the process service in the enterprise informatization age, it is also necessary to integrate data intelligence in the operation management process, and support the real-time control and lean operation of operators' 6G services through data-driven intelligent converged services. Therefore, in the design of PDCA cycle architecture for operation management of 6G BSS, it is required to support the domain-based workflow in various domains horizontally and realize PDCA closed-loop management in various domains driven by digital intelligence; it is required to realize the integration and coordination of various domains vertically in different links such as PDCA, integrate the operation management and control of various domains from the perspective of the overall strategy of enterprises, empower enterprises with global perception and real-time control, and support the continuous circulation and improvement of operation management via the convergence of process and data intelligence, so as to realize the refined operation management and control of enterprises. Specifically:

- **Horizontally, it is required to realize the closed-loop process from plan, do, check and act in the four major domains of resource, production, business and ecosystem:**

  Operators need 6G BSS to realize the digital operation management architecture from operation goal, planning, business implementation, business performance and management control in different domains. At the strategic level, the whole business activities form a complete PDCA management cycle from the enterprise operation plan of 3~5 years, the annual business plan, the production implementation plan to the actual purchase, sales and production order implementation, as well as the management control, performance analysis, problem improvement and implementation adjustment in the business implementation process.

- **Vertically, it is required to realize the integration and collaboration throughout all domains and links through the operation domain and the experience domain:**

  PDCA architecture provides a brand-new digital intelligent enterprise operation management mode, and every business economic activity in each functional domain can be transmitted and converted into data description at the level of operation control and experience management in real time, thus empowering enterprises with global perception and real-time control in operation management and total experience.

Compared with TM Forum ODA functional architecture, 6G BSS, a new generation of business innovation platform based on PDCA, emphasizes the introduction of digital intelligence means for analysis and check in the process management after defining strategic planning and operation implementation in eTOM business architecture, and takes appropriate actions to improve the planning implementation process. This change in management operation domain will promote the rapid iteration and continuous innovation of the 6G digitalization business. Learning from the mode of Internet vendors' opening IaaS and PaaS services in the work domain, 6G BSS decomposes ODA's Production into resource domain and production domain, operates computing network resources based on 6G network in a more open way, and further integrates DOICT to create corresponding production capability. At the same time, the evolution of the business model oriented to the 6G era upgrades ODA's Party Management to the ecosystem domain, which is no longer limited to the management of cooperative participants, but more systematic management of rights and incentive methods, and establishes a digital ecological collaboration support system with operators as the core. In addition, the total experience through the whole digital support system will become the key to establishing customer-centric sustainable development in the 6G era, and it is necessary to manage the digitalization of all participants in the ecosystem by means of digital intelligence.

## 6.2 6G BSS Architecture Overview

To sum up, in order to meet the needs of convergence and evolution of 6G key business scenarios and DOICT, 6G BSS architecture will be designed with the center on "digital intelligent operation capability and domain process optimization". Through the evolution of architecture, 6G BSS will efficiently support the ecological construction of operators in the 6G era, and improve customer business experience satisfaction and agility of digital business innovation.

First of all, 6G BSS needs to design a closed-loop operation process in the four major domains of ecosystem, business, production and resource from top to bottom according to the future fully open and ecological business model and PDCA architecture:

- **Ecosystem domain:** In the 6G era, the first thing operators need to consider is what kind of business ecosystem to establish and how to maintain it through digital operation means. Therefore, 6G BSS shall provide suitable value orientation, collaboration mode and operation mode for different partners, such as traditional supply chain mode, operator-centered platform empowerment mode, and decentralized innovation mode established through rights and drives. Although the above modes are different in connotation and form, how to realize the closed loop of operation process can be unified according to PDCA architecture. Based on PDCA's definition of the operation process, 6G BSS shall realize value planning, ecological collaborative innovation interaction, ecological operating benefit analysis and corresponding rights and drives under different modes, and constantly iterate and optimize. In terms of function, the main functions of the 6G BSS ecosystem domain include new partner management, customer intention management, trusted transaction management, multi-party intelligent bargaining management, smart contract management, ecological value analysis and the like. In terms of ecological construction, in addition to the traditional supply chain ecosystem for communication operators and the cloud computing service ecosystem established in the 5G era, it is necessary to pay more attention to the data value service ecosystem, metaverse digital asset transaction ecosystem and content creator ecosystem construction in the form of decentralized organization.

- **Business domain:** After establishing the enterprise value orientation and the corresponding business ecosystem, aiming at key markets and based on key products, operators and partners will complete the digital intelligent closed-loop operation of relevant market product planning, product ordering and sales, customer operation analysis and supply & marketing strategy adjustment through the digital support system provided by 6G BSS. In the business domain, the main functions of 6G BSS include dimension management, measurement management, product life-cycle management, commodity dynamic portfolio management and customer operation analysis, etc. In the 6G era, operators can support more diverse products and services, including not only enhanced holographic communication services, immersive experience services, Joint Communications and Sensing services, air-space-ground joint communication services, but also distributed intelligent services, twin aggregate services, zero-trust secure and trusted services, brain-computer communication services and so on. However, it is difficult for operators to provide full-stack solutions only with their products and services, whether for consumers or industrial enterprises. Therefore, in order to provide competitive solutions by strengthening ecological cooperation, 6G BSS must be more open and can effectively support closed-loop business operations for different business models.

- **Production domain:** Products and services must be built on efficient and agile production capability. In the 6G network design, the trend of technology convergence is strengthened, creating technical and commercial conditions for building an operator-centered development ecosystem. For PDCA architecture design, operators can build an open developer ecosystem with the support of BSS functions in technical capability planning, capability development and deployment, capability utilization and analysis and capability upgrade and maintenance. Depending on the ecological collaboration of developers, a variety of DOICT bearing business requirements can be transformed into native technical components of distributed cloud, and provide high-quality and low-cost production capability for business innovation via assembled architecture. The main functions of the production domain include cloud-native technology component management, integrated development management and related developer-oriented toolkit management. Oriented to the convergence trend of digital-real under Web3.0 and the accompanying comprehensive integration capability of technologies, 6G BSS will become an integrator of technology integration, including CT technologies such as information perception and semantic transmission brought by 6G network, digital native IT technologies such as XR for metaverse, OT gradually maturing based on digital twin in physical industry, and data intelligent DT in rapid development.

- **Resource domain:** 6G network strengthens operators' capability to control the infrastructure resources. On the one hand, via co-construction, sharing and interconnection and intercommunication, operators expand the resource base of connection services, including not only network resources such as ultra wireless broadband, ultra-large-scale connection, local private network, space-air-ground integrated communication, but also connection service sources such as intelligent perception and high-accuracy positioning. On the one hand, facing the demand for integrated basic resources of digitalization services, operators further improve the layout of the resource domain through self-construction or cooperation with cloud vendors, relying on technologies such as Computing-Aware Networks. Based on PDCA architecture, 6G BSS provides cycle process support for basic resource planning, resource procurement operation, operational efficiency analysis and resource optimization and adjustment in the resource domain. In terms of functions, the main functions include resource scheduling management, resource life-cycle management, resource access authentication service management, task-based resource metrics management and the analysis & optimization of overall resource utilization. Computing-Aware Network will be an integrated basic resource that operators focus on building, in which different heterogeneous computing powers will evolve with the development of services.

Secondly, in addition to supporting the closed loop of operation management process in key domains, 6G BSS needs an enterprise-level operation management center to coordinate various domains to complete integrated collaboration, control and optimization. This White Paper positions the operation domain vertically distributed as the key support system of enterprise strategic management.

- **Operation domain:** In the 6G era, operators can further connect B and M domains through digital and intelligent convergence of business and finance, and establish a full-link operation process supporting enterprise operation management. According to PDCA mode, 6G BSS can set a closed-loop process of enterprise strategic planning, domain-based strategic driving, operating benefit analysis and operating problem handling in the operation domain. Among them, after formulating the overall strategic direction and financial objectives, the enterprise strategic planning is decomposed into ecosystem domain, business domain, production domain and resource domain to complete the detailed planning further, and then judge and make decisions by summarizing the planning contents of various domains. The domain-based strategic drive provides a digital way for enterprise-level strategic tasks to drive real-time collaboration and efficient implementation in related domains. In this process, all interactive information and data will be recorded in the support system of the operation domain. Operating benefit analysis will be carried out and predicted by intelligent method against KPI target using operation data provided by enterprise-level big data platform. Feasible treatment solutions will be recommended as much as possible, which will be improved at the operational problem handling link. In the 6G era, enterprise operations will be more ecological and open, so 6G BSS may need to make intelligent strategic decisions and optimization not only according to the internal data of enterprises, but also concerning the operational data of ecological partners and competitors.

Finally, 6G BSS needs to establish an enterprise-level experience management center to globally perceive and continuously improve the digital experience of customers, partners and employees, so that the experience domain will provide stakeholders in the above domains with the capability to evaluate and improve the digital experience.

- **Experience domain:** 6G BSS needs to strengthen the total experience and accelerate the upgrade of multi-mode interactive experience brought by the 6G network. On the one hand, the experience domain needs to enhance the management of the total experience for customers, employees and users through new triggers such as customer avatars; on the other hand, it also needs to explore the integrated experience brought by intelligent terminals, such as the robot, XR human-computer interaction, brain-computer interface interaction and other multi-mode interaction capabilities. The main functions of experience operation include experience content loading and unloading management, intelligent terminal parameter management, experience data collection management, experience content distribution management and service subscription management.

Overall, 6G BSS fully supports enterprises to complete top-level strategic planning, domain-based efficient collaborative implementation, and upgrade operational experience through digital intelligence on an open architecture (as shown in *Figure 6-2*). On the one hand, 6G BSS will further optimize the process oriented to the future, establish a domain-based KPI system and smooth automated closed-loop operation; on the other hand, it also strengthens the analysis and security check by means of digital intelligence, and takes appropriate actions to improve the planning implementation process.

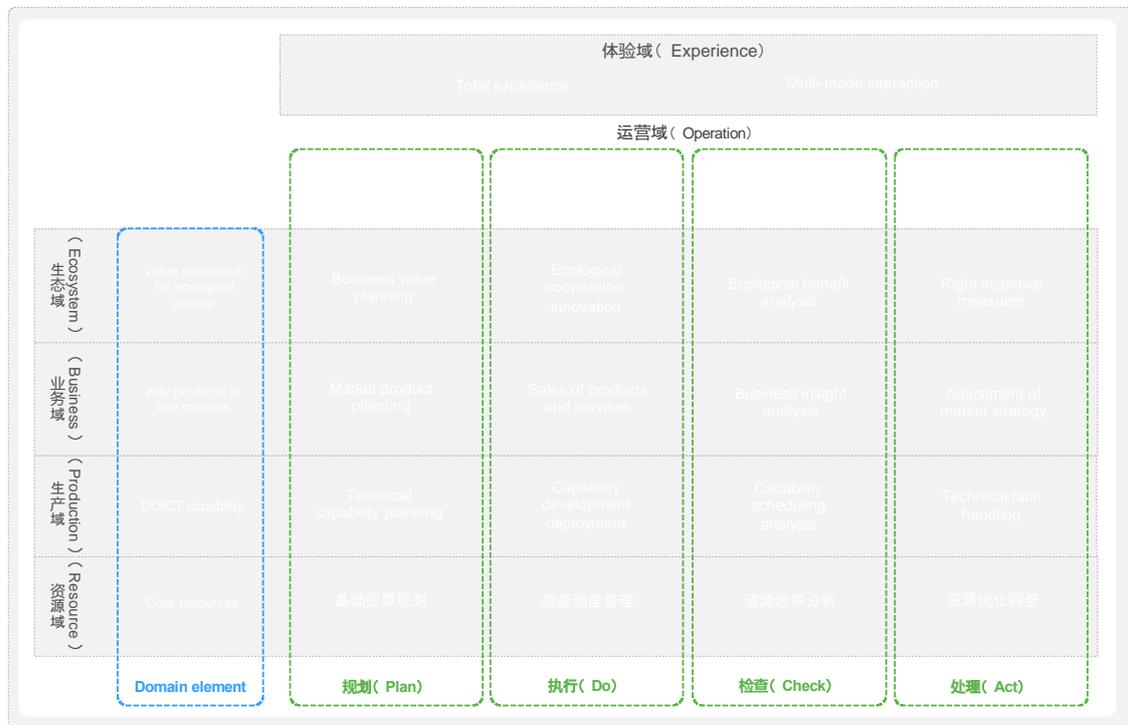

**Figure 6-2 6G BSS Reference Architecture**

At the same time, via virtualization hardware, multi-tenant architecture software and domain-based construction, resource sharing in a wider range can provide operators with lower cost and better experience services, realize BSS migration to the cloud, and meet the maximum intensification and maximum efficiency of industrial value chain. Cloudified 6G BSS is an evolved BSS system, which can better solve the goals difficult to achieve by traditional BSS such as providing better service quality and customer experience with lower construction and operation & maintenance costs, and adapting to more flexible business model changes. For example, an SLA-based self-healing mechanism can speed up fault handling and improve customer experience; full sharing of resources can reduce the cost of hardware procurement; intelligent automated management can reduce operation & maintenance costs and the like.

Therefore, these functional upgrades of 6G BSS will not only greatly enhance the internal collaboration efficiency in various domains, but also effectively support external ecological partners to participate in ecological construction in almost all domains in an open form.

.

# VII. 6G BSS Engineering Principles and Recommendations

4G is a watershed in the development of BSS for operators. 4G and its previous services mainly provide connection channels for C-terminal users. The core of services remains unchanged, but the contents carried on the channels are expanding from voice, messages and texts to images and data. BSS for 4G is designed to support standardized product operation. It forms a structure with CRM and BOSS as the central support systems, focusing on internal supply chain system management and product operation business process streaming.

Taking the Internet of Everything as its design goal, 5G is committed to providing communication services such as massive and reliable connection for industry scenarios. The customer group of operators has been greatly expanded, and the 2B industry has become a breakthrough domain for operators to achieve growth. At the same time, the traditional BSS system, oriented to the 2C industry and supporting standardized product operation, is facing challenges. To this end, during 5G BSS (as shown in *Figure 7-1*) construction, on the one hand, the system introduced the central platform system of cloud computing, integrated resources, middleware, data, business components, etc., and provided composable modules for thousands of businesses to meet the differentiated needs of 2B customers efficiently; on the other hand, BSS began to gradually integrate the capability of OSS internally, and open up to introduce partners' capabilities externally to support operators to carry out personalized ICT converged services.

As a general-purpose technology, 6G strengthens the vision of value spillover and is committed to promoting the self-creation and self-evolution of social and economic values. Therefore, on the basis of deepening the digital operation of its own business, 6G BSS will pay more attention to building the supporting capability of ecological operation, forming a business innovation platform system, and promoting capability exposure, technology convergence and ecological cooperation.

# 7.1 5G BSS Architecture and Extensibility

5G BSS is an essential change for operators from standard product operation to the personalized service operation, laying a specific technical and commercial foundation for 6G BSS.

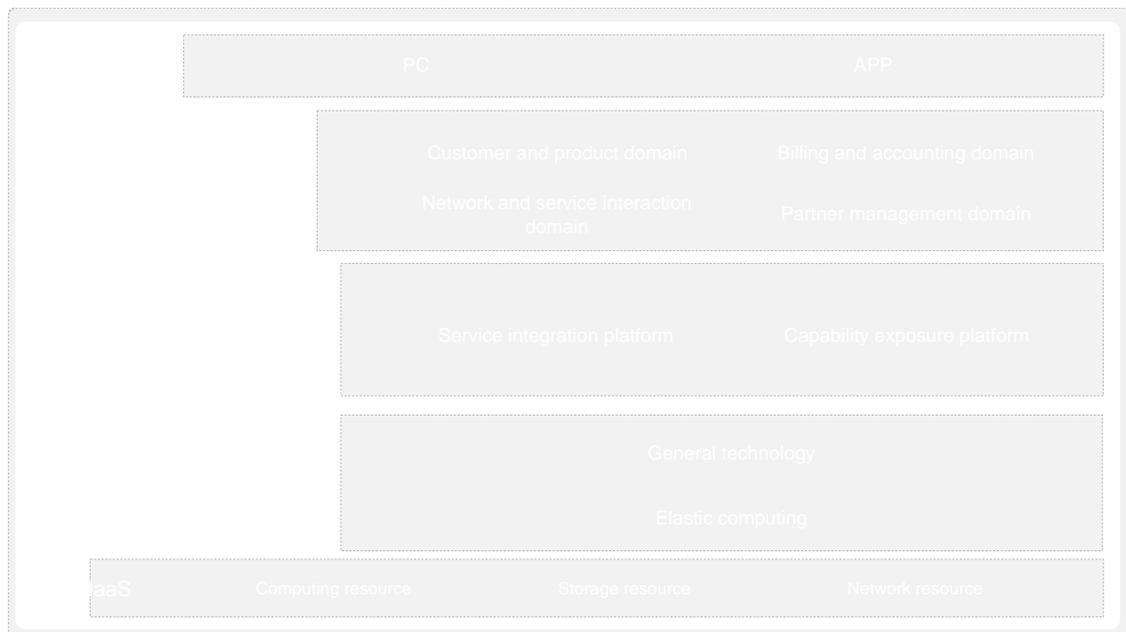

**Figure 7-1 Status of 5G BSS Five-tier Architecture**

*Figure 7-1* shows a certain operator's BSS status and a typical BSS architecture for 5G[56, 57]. Through the analysis of *Figure 7-1*, 5G BSS presents the following characteristics, which build the foundation for the evolution of 6G BSS:

- Introducing cloud computing technology system, forming a support system of hierarchical decoupling and modular delivery

  a. Using cloud computing technology, change how device resources are purchased and deployed according to systems and decouple business and infrastructure resources.

  b. 5G BSS expands the PaaS layer in the three-tier service architecture of cloud computing, forming I-PaaS and T-PaaS, and accumulating the capabilities of service integration capability components and IT technical standard components related to BSS side.

  c. Provide standardized capability opening and integration capabilities, and empower business innovation based on standardized capability assets and T-PAAS technology products.

- Domain-based design of business functions

  5G BSS decomposes the business center into customer and product domain, billing and accounting domain, network and service interaction data domain and partner data domain. The customer and product domain provides the customer full life cycle management functions, including the customer center, order center, products/commodities center, service center, marketing center and so on. The billing and accounting domain provides the functions of online message/offline telephone bill collection preprocessing, batch billing, accounting processing and management, comprehensive settlement and so on. The network and service interaction data domain contains the collection of related business and management data information owned by the domain, including acceptance center and network policy management. The partner data domain contains the process and data support required by the domain application.

- Communication services extending to integrated ICT services

  5G BSS not only considers the support of communication services, but also provides personalized service support for customers, including:

  ➢ Service support based on 5G network capability exposure, such as QoS, slicing and private network.
  ➢ CHBN convergence support for individuals, families, groups and new businesses include 5G personal communication and home broadband convergence and bundled services, 5G industry dedicated line services, and various rights sharing services.

  ➢ Cloud and network converged services, for example, service support capability of public cloud provided by operators, convergence and bundled services of network products and cloud products, and opening of cloud and network convergence.

## 7.2 Recommendations for Evolution towards 6G BSS

In the process of moving from 5G to 6G, BSS will fully absorb the existing architecture, technical system and business model design, fully adopt the practice and exploration achievements of the 5G+ era, and build a hierarchical evolution methodology.

### 7.2.1 Business-driven Gradual Evolution

Technological innovation drives the ever-changing business, which promotes the change and evolution of the BSS. Only BSS that follows the business requirements and leads the way appropriately can have excellent business support and monetization capabilities. It can simultaneously release the flexibility of the huge system and promote the rapid development of the business. Although the industry has made good explorations on the application scenarios, key technologies and business models of 6G and a forward-looking summary of 6G business support has been made, which is of excellent guidance, 6G is far from reaching the phase of maturity in implementation for all technologies and businesses. The business support update in the exploration period shall be gradually promoted according to the needs and trends of the business, fully considering the stable operation support of existing businesses and gradually providing forward-looking support for new business capabilities at the same time.

### 7.2.2 Evolution Based on Open Architecture

In the 6G era, the business support of communication industry needs to maintain good business and technical exposure. Regarding business, it is necessary to provide openness to branches at all levels, industry customers and external partners, and realize the support for business innovation. Maintain good technical compatibility to adapt to the emerging new technologies. On the basis of keeping the architecture openness, it is necessary to further complete the ecosystem construction and form more monetization channels. It is necessary to fully consider flexible performance support, tenant business data isolation, system security, flexible combination, intelligent management and control, etc.

### 7.2.3 Full Migration to Public Cloud/on-Premise Cloud

With the continuous development of information technology, the mobile communication industry has constantly undergone innovations and upgrades. As an essential part of the mobile communication industry, 6G BSS of mobile communication network is always improving and perfecting its function and performance. For 6G BSS, the traditional deployment usually requires a large number of hardware devices and software platforms, and needs to invest a lot of human resources for maintenance and management, which is costly and complex to manage. To solve these problems, migrating 6G BSS to a public cloud or on-premise cloud may be considered.

Migration of 6G BSS to a cloud platform can reduce hardware and software costs and human resources investment, thus reducing the costs. Public cloud and on-premise cloud platforms are provided with flexible scalability, and can quickly allocate and release resources according to business requirements, thus improving flexibility and scalability. Cloud platforms usually are provided with high availability and fault tolerance, which can automatically implement fault tolerance and recovery when hardware or software fails, thus improving the stability and availability of the system. In addition, cloud platforms usually have better data security, providing various security measures which can better protect the security of sensitive data. Fully migrating 6G BSS to the cloud platforms can realize centralized management and monitoring, thus reducing management complexity and workload. At present, 90% of foreign operators are migrating to the public cloud, and most domestic operators in China are also implementing or preparing to implement on-premise cloud and public cloud migration solutions.

The full migration of 6G BSS to the public cloud or on-premise cloud can bring many advantages, including reducing costs, improving flexibility and scalability, improving stability and availability, improving data security, simplifying management and so on. These advantages can help operators better meet business requirements and enhance their competitiveness and innovation capabilities. Therefore, it has become a trend and inevitable choice to fully migrate 6G BSS to a public cloud or on-premise cloud.

### 7.2.4 Native Network Security System

According to the typical application scenarios of 6G, different from previous communication technologies and 5G, the definition of "person" in 6G is further generalized to form a "body area network", where the entities have the same identity to a certain extent, but different signs; even if these entities have the same identity, when facing different roles, their security and authorities shall be different. For example, the user is the owner, and he/she shall have all the required authorities and capabilities. When facing the device provider, it is entirely different, involving the owner's privacy, so the differences in security and authorities are significant. The further intelligent connection between the person, physical world and digital world will involve more physical and logical device entities. So it is essential to know how to manage the security between these entities, persons and services. The security of the Internet of Everything is different from the previous ones, such as inputting user name, password, and fingerprint. Because this kind of security involves many devices and services, oriented to different users, different services and scenarios and the devices shall operate independently and may switch identities at any time, the security requirements are much more complicated. Based on technologies such as Native Network Security mechanism established from native intelligence and zero trust, 6G builds the security base of the whole BSS system. Native Network Security is based on the real-time state of trustworthy relationship, behavior and device scenario. It provides trust relationship, risk management, and control with AI's help and a safer, more flexible and reliable zero-trust security model.

## 7.2.5 Independent Evolution of Interaction Layer

With the rapid development of science and technology, human-computer interaction technology is developing day and day rapidly from text console to CS single application, from Web1.0 and Web2.0 to mobile Internet app and applet, from 5G XR to 6G holographic communication, body area network wearing and brain-computer interface, etc. It is vital to separate the front-end interaction part from the service support part in BSS system to ensure the independent evolution of the interaction layer. It will significantly improve the business experience, meet the hierarchical service requirements at different levels and expand the business market scope if the rapid upgrade of the interaction layer of the front end is independently supported. Interaction layer independence is not only a "separation of front and back ends", but also a matter of operation & maintenance issues such as unified access, unified portal, low code platform, zero-trust Native Network Security and industry application scenario support and DevOps to be considered.

## 7.2.6 Synchronization of Business, Data and Intelligence

6G businesses are more complex than those before, which involves not only all kinds of clouds, networks, edges, computing, industries, security, intelligence, etc., but also more new wearable devices, industrial control devices, space-air devices, etc. provided as well as new models such as body area network, new dynamic intelligent price management, new intention recognition, a large number of perception, timely service and other capabilities provided. These complicated capabilities will further promote the closer collaboration and synchronous evolution of the trinity of business, data and intelligent platforms. On the one hand, in the face of the complexity of business, from the perspective of marketing, business opportunities, intentions, operation & maintenance, it is inseparable from the assistance of AI. Only by making full use of the capability of AI can automation, intelligentization, cost reduction and efficiency improvement, could shift peak load and providing reasonable solutions be achieved; on the one hand, a large number of businesses offer massive amounts of data, and due to the complexity of 6G business, the knowledge relationship between data is superimposed more intensively. Only by making full use of big data technology as the foundation can a sound data foundation guarantee for business and AI be provided, otherwise 6G intelligence will be empty talk; on the other hand, AI will become more mature in 6G businesses, such as 6G dynamic pricing, intention perception and recognition, recommendation of integrated solutions, and QoE of perceived data convergence, which will provide unprecedented opportunities for the development of AI.

## 7.2.7 Rapidly Empower Enterprise Customer through Ecosystem

6G is more oriented to toC businesses on the basis of 5G B2B2X, while for 6G empowered vertical industries oriented to toC, toH, toF and other businesses, it shall not only consider 6G technology, business open architecture and ecological construction, at the same time, but also consider the rapidly empowered vertical industries' own business ecosystem construction and rapid business refinement, including ecosystems that are not limited to technology, business, operation, management, cooperation, self-service, etc.; because vertical industries are subject to requirements in business security, user security, data security or security regulations, etc., on the one hand, it is necessary to consider adopting refining methods similar to 5G UPF, SMF, etc. to rapidly deploy to the edge nodes of the park to ensure that data will not leave the park and the data are isolated, etc.; on the other hand, it is also necessary to help vertical industries intelligently solve operation and operation & maintenance problems via federated learning, so as to help vertical industries reduce costs and increase efficiency. Only by assisting vertical industries to support and develop business quickly can 6G business bloom.

# VIII. Summary and Prospect

As the first forward-looking and systematic white paper on 6G BSS in the world, after reviewing the development history of BSS in different generations, combined with the overlapping development characteristics of communication businesses and Internet businesses, based on BSS reference architecture in the communication domain and the business model in the Internet industry, this White Paper puts forward the 6G-oriented domain-based business support model and the development vision of 6G BSS. In this White Paper, the characteristics of 6G BSS integrated communication business and Internet business in the future will gradually step out of the double closed-loop systems of "internal support" and "external empowerment", releasing more business innovation and business model innovation for the industry. Based on this prediction, this White Paper divides the core functions of 6G BSS into four domains: resource, production, business and ecosystem, with two domains of operation and experience applied through the above four core functions. Finally, based on the methodology system of PDCA, this White Paper presents the functional architecture and thirteen potential key technologies of BSS for 6G.

During the continuous transformation of operators' BSS for the future, 5G BSS has provided a good development foundation for 6G, including the evolution of cloud computing technology systems and the scalability of service-based communication capability, etc. However, the non-standard, personalized and ecological win-win operation mode for the 2B industry still needs further exploration and iteration. Therefore, based on the goal of 6G commercialization in 2030, this White Paper makes the following prospects for the development phases of 6G BSS:

In 2025, the characteristics and technical system of 5G BSS will be inherited and iterated and the sharing capability of the resource domain and the flexibility capability of the production domain will be realized first; a preliminary exploration will be made on the scenario-based experience capability and operation streaming; related vital technologies such as interactive mode upgrade based on intention perception, intelligent customer avatar, visual operation based on intelligent digital twins, distributed application based on cloud-edge-terminal integration, PaaS capability based on sustainable cloud computing model and cloudified architecture based on high performance will be adopted in priority.

In 2027, the capability oriented to the flexibility of the production domain will be further evolved; minimalist development will be provided based on super automation and platform engineering, and related trusted identity authentication and Native Network Security capabilities will be further strengthened; in terms of diversified capability building of business domain, the construction of joint communication, sensing and computing service and air-space-land integration service primarily provided by 6G will be supported and QoPE operation mode and new trading capability based on blockchain will be primarily explored.

In 2030, in the middle and late phases of 6G development, with the iterative evolution of business models, some cooperation paradigms will be adopted, the valuation capability of the ecosystem domain will be further evolved in BSS, and new business models will be piloted; the key products and services in the business domain will be supported by the full life cycle system combined with the business model design in the ecosystem domain to achieve the internal operation digitalization promotion and the external value flow closed loop. At this phase, the ecological valuation of data exchange capability and automated capability of operation & maintenance need to be enhanced, and related technologies such as data capitalization based on data fabric, data value maximization based on privacy computing, automated operation based on self-adaptive AI, and automated operation & maintenance based on digital immune system will be fully implemented.